\documentclass{article}

\PassOptionsToPackage{numbers,compress}{natbib}
\usepackage[preprint]{neurips_2026}

\usepackage[utf8]{inputenc}
\usepackage[T1]{fontenc}
\usepackage{hyperref}
\usepackage{url}
\usepackage{booktabs}
\usepackage{amsfonts}
\usepackage{amsmath}
\usepackage{nicefrac}
\usepackage{xcolor}
\usepackage{graphicx}
\usepackage{multirow}
\usepackage{array}
\usepackage{enumitem}
\usepackage{amsthm}
\usepackage{pifont}
\usepackage{subcaption}
\usepackage[most]{tcolorbox}
\usepackage{tikz}
\usetikzlibrary{arrows.meta, positioning}

\newtcolorbox{promptblock}[1]{
  enhanced,
  colback=white,
  colframe=black,
  colbacktitle=black,
  coltitle=white,
  fonttitle=\bfseries\small,
  title={#1},
  boxrule=0.6pt,
  arc=2pt,
  left=4pt,right=4pt,top=3pt,bottom=3pt,
  before skip=4pt, after skip=4pt,
}

\newcommand{\bench}{\textsc{GraphInfer-Bench}}

\newcommand{\masked}{\texttt{[MASK]}}
\newcommand{\eg}{\emph{e.g.},\ }

\title{\bench: Benchmarking LLM's Inference Capability on Graphs}
\author{%
  Zhuoyi Peng$^{1}$ \quad Jingzhou Jiang$^{1}$ \quad Hanlin Gu$^{2}$ \quad Lixin Fan$^{2}$ \quad Yi Yang$^{1}$ \\
  $^{1}$The Hong Kong University of Science and Technology \quad $^{2}$Webank
}

\begin{document}
\maketitle

\begin{abstract}
Graph analysis underlies many applications whose answers cannot be
looked up in a single record or retrieved along a path: laundering
rings, drug repurposing, user preference, and scientific theme are
all inferred from a node together with its neighbourhood. We
introduce \bench, a benchmark for whether LLMs can perform this
\emph{graph inference}: producing an open-ended answer that no
single node supports and no path retrieves. Existing graph-QA
protocols cannot test this capability: algorithm simulation, node
classification, single-node description, KG-QA, and GraphRAG all
admit answers retrievable from one node or along a path.
\bench\ defines five tasks along \emph{Description} (what a region
is) and \emph{Comparison} (how regions differ), each constructed
so the ground truth lives in no single node. The release contains
42{,}000 samples across six real-world graphs, produced
automatically and screened by a four-layer quality-control
protocol. We evaluate four method families against the same tasks:
graph-token alignment models, zero-shot frontier closed-source
LLMs, Graph2Text supervised fine-tuning, and plain GNNs as a
structural reference. No method family closes the gap. Graph-token
alignment partially handles description tasks (relational, theme)
but collapses on comparison tasks. Frontier LLMs lead on outlier
detection and community partition among LLM-based methods but lag
on masked-node prediction. Graph2Text SFT is the strongest
LLM-based method on the description side yet falls behind frontier
LLMs on comparison. Across every task, plain GNNs match or beat the
strongest LLM-based row, with the largest margin on community
detection. \bench\ surfaces graph inference as an open capability
gap rather than a property of any one architecture.

\smallskip\noindent\footnotesize
Code: \url{https://github.com/graphinfer/GraphInfer-Bench}.\hspace{0.6em}%
Dataset: \url{https://huggingface.co/datasets/graphinfer/graphinfer}.
\end{abstract}

\section{Introduction}
\label{sec:intro}

Graph analysis underlies many real-world problems whose answers
cannot be looked up in a single record or retrieved along a
path. A money-laundering ring is identified by a pattern of
transactions across many
accounts~\cite{motie2024financial}. A drug repurposing
hypothesis emerges from joint reasoning over drug-gene-disease
relationships~\cite{himmelstein2017systematic}. A user's preference is
inferred from the structure of past interactions rather than from
any individual purchase~\cite{wu2022graph}. A scientific theme is
read off the joint citation pattern of many papers, not from any
one abstract~\cite{newman2012communities}. None of these answers
exists in any single node: each must be \emph{inferred} from a
node together with its neighbourhood.

\textbf{Graph inference.} We define \emph{graph inference} as
producing an open-ended answer to a question about a node and its
neighbourhood, where the answer
(i) is undetermined by any single node's content,
(ii) is undetermined by any traversal value (excluding KG-QA-style
retrieval), and
(iii) requires jointly reading edges and node text rather than
either modality alone.
\bench\ measures the capability, not any particular architecture.

\textbf{Graph inference is distinct from lookup and retrieval.}
\emph{Lookup} returns a node's own attribute. \emph{Retrieval}
returns a pre-existing answer reached through the graph (a KG-path
entity, a span from one node's page). \emph{Inference} is neither:
identifying a cluster's unifying theme, the outlier that does not
belong, or a coherent partition all require open-ended language
that exists in no single node and is not retrievable along any
path.

\textbf{Existing graph-QA evaluations do not test it.}
The protocols in Table~\ref{tab:eval_gap} are lookup or retrieval,
not inference. Algorithm simulation (NLGraph, GraphArena, and
related) is structural lookup over synthetic graphs. Node
classification, link prediction, and single-node description
(LLaGA) are single-node tasks. KG-QA (WebQSP, CWQ, GrailQA,
MetaQA, KQA Pro) is path-traversal retrieval. GraphRAG (STaRK,
CRAG, GRBench) is corpus retrieval indexed by a graph. \bench\
asks for themes, outliers, partitions, and masked content over
six real-world graphs, scored against deterministic structural
ground truth.

\begin{table*}[!t]
\vspace{-3\baselineskip}
\centering
\caption{Graph-QA evaluations in current use, organised along
  the axes \textbf{G} graph, \textbf{Q} question, \textbf{A}
  answer.}
\label{tab:eval_gap}
\vspace{2pt}
\footnotesize
\renewcommand{\arraystretch}{1.2}
\setlength{\tabcolsep}{4pt}
\resizebox{\textwidth}{!}{%
\begin{tabular}{@{}p{4.5cm}p{2.6cm}p{3.2cm}p{3.4cm}@{}}
\toprule
\textbf{Task}
  & \textbf{G: graph}
  & \textbf{Q: question}
  & \textbf{A: answer} \\
\midrule
\multicolumn{4}{c}{\textbf{Structure Lookup}: answer from graph structure alone} \\
\midrule
\textbf{Algorithm simulation}\newline {\scriptsize NLGraph~\cite{wang2023can}, GraCoRe~\cite{yuan2025gracore}, GraphArena~\cite{tang2024grapharena}, Talk-Like-a-Graph~\cite{fatemi2023talk}, GraphWiz~\cite{chen2024graphwiz}}
  & \textbf{Synthetic} \newline {\scriptsize \emph{e.g.,} ER / SBM random graph}
  & \textbf{Connectivity, shortest path, cycle, degree, topological sort} \newline {\scriptsize \emph{e.g.,} ``shortest path $v_3 \to v_7$?''}
  & \textbf{Number, yes/no, path} \newline {\scriptsize \emph{e.g.,} \texttt{4} or \texttt{[v3,v5,v7]}} \\
\hline
\multicolumn{4}{c}{\textbf{Retrieval}: answer retrieved from content stored in the graph} \\
\midrule
\textbf{KG-QA}\newline {\scriptsize WebQSP~\cite{yih2016value}, CWQ~\cite{talmor2018web}, GrailQA~\cite{gu2021beyond}, MetaQA~\cite{zhang2018variational}, KQA~Pro~\cite{cao2022kqa}}
  & \textbf{Real KG} \newline {\scriptsize \emph{e.g.,} Freebase, Wikidata}
  & \textbf{Multi-hop lookup} \newline {\scriptsize \emph{e.g.,} ``actors in films directed by Nolan?''}
  & \textbf{Entity / number} \newline {\scriptsize \emph{e.g.,} \texttt{\{DiCaprio, Bale\}}} \\[10pt]
\textbf{GraphRAG / retrieval-on-graph}\newline {\scriptsize STaRK~\cite{wu2024stark}, CRAG~\cite{yang2024crag}, GRBench~\cite{jin2024graph}}
  & \textbf{Corpus + index} \newline {\scriptsize \emph{e.g.,} STaRK-PrimeKG}
  & \textbf{Retrieve then answer} \newline {\scriptsize \emph{e.g.,} ``side effects of drug $d$?''}
  & \textbf{Entity / Factoid / Span} \newline {\scriptsize \emph{e.g.,} span from $d$'s page} \\
\hline
\multicolumn{4}{c}{\textbf{Weak Inference}: a single node (or endpoint pair) suffices} \\
\midrule
\textbf{Node classification}\newline {\scriptsize OFA~\cite{liu2023one}, GLBench~\cite{li2024glbench}, GraphGPT~\cite{tang2024graphgpt}}
  & \textbf{Real TAG} \newline {\scriptsize \emph{e.g.,} ogbn-arxiv}
  & \textbf{Pick node label} \newline {\scriptsize \emph{e.g.,} ``category of this paper?''}
  & \textbf{Fixed label} \newline {\scriptsize \emph{e.g.,} \texttt{cs.LG}} \\[10pt]
\textbf{Single-node description}\newline {\scriptsize LLaGA~\cite{chen2024llaga}}
  & \textbf{Real TAG} \newline {\scriptsize \emph{e.g.,} ogbn-arxiv}
  & \textbf{``Describe this paper''} \newline {\scriptsize \emph{e.g.,} ``describe node $v$''}
  & \textbf{Free text on one node} \newline {\scriptsize \emph{e.g.,} paraphrase of $v$'s title} \\[10pt]
\textbf{Link prediction}\newline {\scriptsize OFA~\cite{liu2023one}, LLaGA~\cite{chen2024llaga}, GraphGPT~\cite{tang2024graphgpt}}
  & \textbf{Real TAG / KG} \newline {\scriptsize \emph{e.g.,} Cora, FB15k}
  & \textbf{Edge $(u,v)$?} \newline {\scriptsize \emph{e.g.,} ``does $u$ cite $v$?''}
  & \textbf{Yes / no} \newline {\scriptsize \emph{e.g.,} \texttt{yes}} \\
\hline
\multicolumn{4}{c}{\textbf{\bench}} \\
\midrule
\textbf{LLM Inference Over Graph}\newline {\scriptsize \textbf{\bench\ (ours)}}\newline {\scriptsize Task 1: masked node prediction}\newline {\scriptsize Task 2: relational description}\newline {\scriptsize Task 3: theme summarisation}\newline {\scriptsize Task 4: outlier detection}\newline {\scriptsize Task 5: community detection}
  & \textbf{Real TAG}\newline \textbf{(6 domains)} \newline {\scriptsize academic citation, e-commerce, clinical citation, encyclopedia, patent citation, Physics Q\&A}
  & \textbf{Masked node / relation / theme / outlier / community} \newline {\scriptsize \emph{e.g.,} ``partition the 25 papers into research-area communities''}
  & \textbf{Open-ended language} \newline {\scriptsize \emph{e.g.,} \texttt{\{0,3,8\}: Information Theory; \{1,2,5,7\}: Numerical Analysis; \{4,6\}: Computational Engineering.} \emph{Reasoning:} titles in \{0,3,8\} centre on coding and entropy bounds~\ldots} \\
\bottomrule
\multicolumn{4}{@{}l}{\scriptsize TAG: text-attributed graph. \quad KG: knowledge graph.}
\end{tabular}%
}
\renewcommand{\arraystretch}{1.0}
\end{table*}

\paragraph{Contributions.}
\begin{enumerate}[leftmargin=*,topsep=2pt,itemsep=2pt]
  \item \textbf{A benchmark dedicated to graph inference.}
        We give a method-agnostic definition (criteria
        (i)--(iii) above) and build the first benchmark targeting
        this capability rather than any specific architecture,
        distinct from algorithm simulation, node classification,
        single-node description, KG-QA, and GraphRAG.

  \item \textbf{A 42{,}000-sample dataset across six domains and
        five tasks.}
        Six text-attributed graphs span \texttt{ogbn-arxiv},
        \texttt{PubMed}, \texttt{USPTO} patents,
        \texttt{ogbn-products}, \texttt{WikiCS}, and
        \texttt{Physics~SE}. Five tasks span two axes.
        \emph{Description}: T1 masked node prediction (recover a
        held-out node from its neighbourhood), T2 relational
        description (characterise the relation between two
        endpoints), T3 theme summarisation (name the unifying theme
        of an ego-graph). \emph{Comparison}: T4 outlier
        detection (identify the node that does not belong),
        T5 community detection (partition an ego-graph into
        coherent groups). Each sample passes a four-layer quality
        gate (scripted rules, dual 70B judges, human-$\kappa$
        calibration, structural deduplication) at low
        human-annotation cost.

  \item \textbf{Evaluation, results, and what they imply for
        closing the gap.}
        Under matched splits and a unified hard-label plus
        SBERT-cosine protocol, we evaluate four method families
        (graph-token alignment, zero-shot frontier LLMs,
        Graph2Text SFT, plain GNNs as a structural reference).
        \textbf{No family closes the gap.} Plain GNNs match or
        beat the strongest LLM-based row on every task, with the
        largest margin on community detection. \emph{The signal
        is in the graph. Closing the gap is an objective and
        decoding problem, not a capacity problem.}
\end{enumerate}

\section{Related Work}
\label{sec:related}

\paragraph{Existing graph-QA benchmarks.}
We extend the references in Table~\ref{tab:eval_gap}.
\emph{Algorithm simulation}: NLGraph~\cite{wang2023can},
GraCoRe~\cite{yuan2025gracore}, GraphArena~\cite{tang2024grapharena},
Talk-Like-a-Graph~\cite{fatemi2023talk},
GraphWiz~\cite{chen2024graphwiz}, plus
GraphInstruct~\cite{luo2024graphinstruct}.
\emph{KG-QA}: WebQSP~\cite{talmor2018web},
CWQ~\cite{talmor2018web},
GrailQA~\cite{gu2021beyond}, MetaQA~\cite{zhang2018variational},
KQA~Pro~\cite{cao2022kqa}.
\emph{GraphRAG}: STaRK~\cite{wu2024stark}, CRAG~\cite{yang2024crag},
GRBench~\cite{jin2024graph}, plus the two GraphRAGBench
variants~\cite{xiang2025use,xiao2025graphrag}.
\emph{Node classification, single-node description, link
prediction} on real text-attributed graphs:
OFA~\cite{liu2023one}, GLBench~\cite{li2024glbench},
GraphGPT~\cite{tang2024graphgpt}, LLaGA~\cite{chen2024llaga}, plus
GPT4Graph~\cite{guo2023gpt4graph},
G-Retriever~\cite{he2024g}, and the LLMs-as-Predictors /
LLMs-as-Enhancers study of Chen et al.~\cite{chen2024exploring}. Different from these benchmarks, \bench\ the first benchmark to our knowledge that targets
graph inference itself.

\paragraph{Methods for LLM understanding of graphs.}
Two families dominate. \emph{Graph-token alignment} trains a GNN
encoder and projects its output into the LLM input space, freezing
or lightly tuning the LLM. LLaGA~\cite{chen2024llaga},
GraphToken~\cite{perozzi2024let},
GraphGPT~\cite{tang2024graphgpt},
TEA-GLM~\cite{wang2024llms}, RGLM~\cite{zhang2026toward},
GOFA~\cite{kong2024gofa}, and
InstructGLM~\cite{ye2024language} differ in how the projector is
trained (PCA alignment, contrastive, reconstructive, instruction
tuning) but share the architectural commitment that structure
enters the LLM through learned graph tokens.
\emph{Graph2Text} serialises the neighbourhood directly as text and
hands it to the LLM, with the LLM either prompted zero-shot
(frontier closed-source models) or supervised fine-tuned on
graph-to-text targets~\cite{fatemi2023talk,guo2023gpt4graph,chen2024exploring}.
\bench\ evaluates both families on the same tasks alongside plain
GNNs as a structural reference, so the comparison isolates which
attack on the inference capability works where.

\section{Dataset and Task Description}
\label{sec:dataset}

Building on the gap identified above, \bench\ operationalises the
inference test as five tasks over six text-attributed graphs. The
sampler produces 66{,}000 candidates (2{,}200 per task-domain cell).
A four-layer quality-control pipeline filters defective references
and deduplicates the observed graph, after which every cell is capped
to 1{,}400 samples balanced on the gold label and split 1{,}000
train, 100 val, and 300 test. The public release totals
\textbf{42{,}000} samples. Below we describe the raw data sources,
the task taxonomy, and the quality-control pipeline. Per-domain text,
full task setups with example prompts, and verification details are
in the appendix.

\subsection{Raw Data Sources}
\label{sec:raw_sources}

\bench\ is built on six public text-attributed graphs that span the
major graph families used in the GNN-LLM literature
(Table~\ref{tab:domains}): citation networks
(\texttt{ogbn-arxiv}~\cite{hu2020open},
\texttt{PubMed}~\cite{sen2008collective},
\texttt{USPTO}~\cite{leskovec2005graphs}), e-commerce co-purchase
(\texttt{ogbn-products}~\cite{hu2020open}), encyclopedia links
(\texttt{WikiCS}~\cite{mernyei2020wiki}), and a Physics Q\&A graph
(\texttt{Physics~SE}~\cite{stackexchangedump}).
Each domain provides natural-language node text (titles, names,
descriptions) and a structurally distinct edge type.
To construct a (graph, question, answer) sample, we sample 2-hop
ego-subgraphs centred on hub nodes (in-degree $\geq 3$ with valid
text), with up to 10 neighbours per hop.\footnote{This size setting
lets us compare graph-token alignment and Graph2Text baselines
under a single subgraph budget. Larger ego-graphs would saturate
the Graph2Text context window once each node's title is
serialised.}
Models receive node titles only. Abstracts and dataset labels are
withheld for ground truth. Per-domain descriptions appear in
Appendix~\ref{app:domains}.

\begin{table*}[t]
\centering
\caption{\bench\ raw data sources used to construct the
  benchmark, \emph{not} the final released dataset (see
  Sec.~\ref{sec:tasks} for the per-cell sample counts).}
\label{tab:domains}
\footnotesize
\renewcommand{\arraystretch}{1.0}
\setlength{\tabcolsep}{3pt}
\begin{tabular}{@{}p{2.1cm}p{2.3cm}rrrp{2.6cm}p{2.4cm}@{}}
\toprule
\textbf{Domain} & \textbf{Graph type}
  & \textbf{N (nodes)} & \textbf{E (edges)} & \textbf{Labels (categories)}
  & \textbf{Edge meaning} & \textbf{Node text} \\
\midrule
ogbn-arxiv     & Academic citation & 169K & 1.2M & 40 & Paper cites paper   & Paper titles    \\
ogbn-products  & E-commerce        & 2.4M & 124M & 47 & Co-purchased        & Product titles  \\
PubMed         & Clinical citation & 20K  & 44K  & 3  & Paper cites paper   & Paper titles    \\
WikiCS         & Encyclopedia      & 11K  & 276K & 10 & Article links to    & Article titles  \\
USPTO          & Patent citation   & 266K & 2.0M & 19 & Patent cites patent & Patent titles   \\
Physics SE     & Physics Q\&A      & 234K & 105K & 51 & Related questions   & Question titles \\
\bottomrule
\end{tabular}
\end{table*}

\subsection{Tasks}
\label{sec:tasks}

\begin{figure*}[t]
\centering
\begin{tikzpicture}[
    every node/.style = {rounded corners=2pt, draw, align=center,
                          inner sep=2pt, font=\tiny},
    arrow/.style = {-Latex, gray!70, line width=0.3pt},
    root/.style  = {fill=blue!8,  draw=blue!50,  thick, font=\scriptsize\bfseries},
    fam/.style   = {fill=teal!10, draw=teal!50,  thick, font=\scriptsize\bfseries,
                    minimum width=38mm},
    leaf/.style  = {fill=white,   draw=black!50, minimum width=23mm,
                    text width=23mm, font=\tiny}]
  \node[root] (root) at (0,0) {Graph Inference Tasks};
  \node[fam] (desc) at (-3.0, -1.0) {Description \\
        \scriptsize\mdseries\itshape describe node, edge, cluster};
  \node[fam] (comp) at ( 3.0, -1.0) {Comparison \\
        \scriptsize\mdseries\itshape compare nodes (outlier, clusters)};
  \draw[arrow] (root) -- (desc);
  \draw[arrow] (root) -- (comp);
  \node[leaf] (t1) at (-5.6, -2.9)
       {\textbf{T1: Masked Node}\\[1pt]
        \textit{Q:} topic of masked Node 0?\\
        \textit{A:} ``Computational Geometry.''\\
        \textit{R:} neighbours cover clustered planarity \& simultaneous embedding.};
  \node[leaf] (t2) at (-3.0, -2.9)
       {\textbf{T2: Relational Sem.}\\[1pt]
        \textit{Q:} does Node 9 cite Node 12?\\
        \textit{A:} ``No.''\\
        \textit{R:} Node 9 cites point-location works, not the working-set tree paper.};
  \node[leaf] (t3) at (-0.4, -2.9)
       {\textbf{T3: Theme Summary}\\[1pt]
        \textit{Q:} theme of this 13-paper graph?\\
        \textit{A:} ``Distributed Computing.''\\
        \textit{R:} titles cite distributed streams, load balancing, parallelism.};
  \draw[arrow] (desc) -- (t1);
  \draw[arrow] (desc) -- (t2);
  \draw[arrow] (desc) -- (t3);
  \node[leaf] (t4) at ( 2.2, -2.9)
       {\textbf{T4: Outlier Detection}\\[1pt]
        \textit{Q:} which paper is unrelated?\\
        \textit{A:} Node 8: ``outlier''.\\
        \textit{R:} Nodes 0--7 are differential-privacy works, Node 8 isn't.};
  \node[leaf] (t5) at ( 4.8, -2.9)
       {\textbf{T5: Community Detection}\\[1pt]
        \textit{Q:} cluster nodes by topic.\\
        \textit{A:} ``Comm.~0: Computer Vision''.\\
        \textit{R:} cluster groups segmentation \& pose-estimation papers.};
  \draw[arrow] (comp) -- (t4);
  \draw[arrow] (comp) -- (t5);
\end{tikzpicture}
\caption{Taxonomy of the five graph inference tasks in \bench.
  \textbf{Description} tasks (T1 through T3) ask about a single target
  node, edge, or local theme. The answer is a single short string
  scored by per-task hard accuracy together with SBERT-F1 against a
  reasoning paragraph. \textbf{Comparison} tasks (T4 and T5) require
  cross-node reasoning over the full ego-graph and produce structured
  outputs: a chosen node for T4, a partition for T5.}
\label{fig:taxonomy}
\end{figure*}
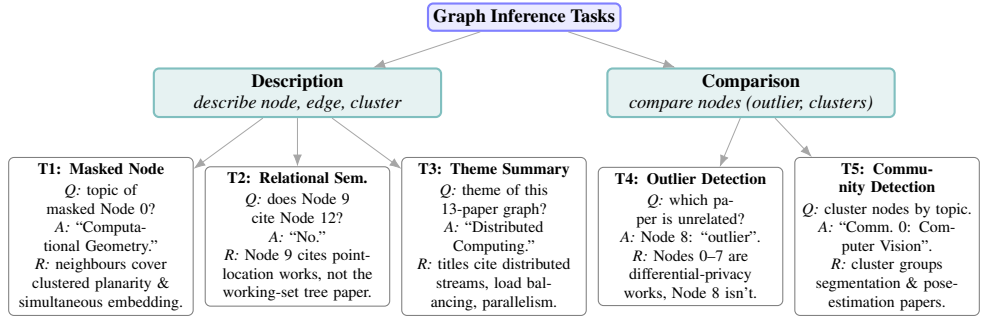

\bench\ organises five tasks under two graph-inference capabilities,
summarised in Table~\ref{tab:tasks}. \textbf{Description (T1 to T3)}
asks the model what a region of the graph \emph{is}.
\textbf{Comparison (T4 and T5)} asks how regions \emph{differ}.
Every task is constructed so that single-node shortcuts cannot
solve it. For T1 the hub's title is masked. For T2 the answer is
split across pair endpoints. Full no-shortcut constructions appear
in Appendix~\ref{app:tasks}.

Every sample is a triple \emph{(input, answer, reasoning)}. The
\emph{reasoning} is a short, structured natural-language
justification that grounds the answer in specific node attributes
and edges. It is auto-generated by an LLM during dataset
construction\footnote{Gold reasoning is produced by the DeepSeek
API as of 2026-04-01. The output is then audited by the four-layer
quality-control pipeline described in Sec.~\ref{sec:quality}.} and passes the four-layer quality-control
pipeline of Sec.~\ref{sec:quality} before release. Models are
trained and scored on both fields jointly: hard metrics on the
\emph{answer}, soft semantic similarity on the \emph{reasoning}
paragraph (Sec.~\ref{sec:metrics}).

\begin{table}[h]
\centering
\caption{The five tasks of \bench, summarised as Question, Answer,
Reasoning, and a real-world application. Full no-shortcut
constructions and per-domain templates are in
Appendix~\ref{app:tasks}.}
\label{tab:tasks}
\scriptsize
\renewcommand{\arraystretch}{1.05}
\setlength{\tabcolsep}{3pt}
\begin{tabular}{@{}p{1.4cm}p{2.7cm}p{2.0cm}p{2.7cm}p{3.6cm}@{}}
\toprule
\textbf{Task} & \textbf{Question} & \textbf{Answer} & \textbf{Reasoning} & \textbf{Application} \\
\midrule
\textbf{T1} Masked node
& Hub's title is \masked. What is the hub's subject label?
& A subject label.
& Cites neighbour titles that pin down the topic.
& Classifying an unlabelled item from labelled neighbours (new arxiv preprint, uncategorised product, unindexed patent). \\
\midrule
\textbf{T2} Relational description
& Given an edge $(u, v)$ inside $u$'s subgraph, does the relation hold, and what role does $v$ play in $u$'s context?
& Yes/no plus a role description.
& Grounded in the local topology around $u$.
& Predicting and explaining link relations (why one paper cites another, why two products co-purchase). \\
\midrule
\textbf{T3} Theme summarisation
& Given a cluster of same-label nodes, what dominant theme unifies them?
& A short theme label.
& Representative titles from the cluster cited as evidence.
& Naming what a graph region covers (citation cluster contribution, patent family technology area, Wikipedia subfield). \\
\midrule
\textbf{T4} Outlier detection
& A coherent cluster plus one structurally embedded but topically distant outlier. Which node is the outlier?
& A per-node label line, with one node tagged as the outlier.
& Contrast between the outlier's topic and the cluster's dominant topic.
& Spotting items that do not belong (off-topic Q\&A post, miscategorised product, miscited paper). \\
\midrule
\textbf{T5} Community detection
& A shuffled graph mixing 2 to 3 categories. How do nodes partition into coherent communities?
& A per-node community id with a one-line name per community.
& Title evidence that grouped each community.
& Recovering latent groupings in mixed neighbourhoods (citation subfields, patent technology areas, product categories). \\
\bottomrule
\end{tabular}
\end{table}

\subsection{Data Quality Control}
\label{sec:quality}

\noindent Every candidate passes through a four-layer release gate.
\textbf{Layers 1 to 3} audit the gold \emph{reasoning} (answer plus
rationale) under a shared five-code vocabulary (Fact, Consistency,
Label-Integrity, Logic, Evidence) so scripted rules, LLM judges,
and human annotators speak the same language. \textbf{Layer 4}
audits the observed \emph{graph} for split leakage.

\begin{itemize}[leftmargin=*,topsep=2pt,itemsep=2pt]
  \item \textbf{L1, agent-grounded regex filter.} A deterministic
        rule set distilled from an agentic LLM audit of the
        candidate pool. Admits $98.98\%$.
  \item \textbf{L2, dual 70B judges with \texttt{all\_pass}.}
        Llama-3.1-70B-AWQ and Qwen-2.5-72B-AWQ vote independently;
        a sample ships only if both mark it qualified. Admits
        $94.0\%$.
  \item \textbf{L3, human calibration of L2.} Two annotators score
        a stratified $300$-sample subset
        independently.\footnote{Each annotator labels every sample
        without seeing the other annotator's verdict, so the two
        label streams are independent before $\kappa$ is computed.}
        We achieve inter-annotator Cohen's $\kappa = 0.606$ and
        L2-vs-human agreement $= 95.2\%$, both above the
        pre-registered threshold.
  \item \textbf{L4, construction-time deduplication.} Per-task
        canonical identities prevent the same ego-centre, edge,
        cluster, outlier pair, or sub-graph from leaking across
        train, val, and test.
\end{itemize}

The pipeline admits $59{,}681 / 66{,}000$ candidates ($90.4\%$),
which we balance and cap to $1{,}400$ samples per (domain, task)
for the released benchmark (Tab.~\ref{tab:stats}). Per-layer rule
sets, prompts, model identifiers, calibration protocol, and
canonical identity definitions are in
Appendix~\ref{app:verification}.

\begin{table}[h]
\centering
\caption{Released sample counts per task per domain, totalling 42{,}000 samples (30{,}000 train, 3{,}000
  val, 9{,}000 test).}
\label{tab:stats}
\footnotesize
\renewcommand{\arraystretch}{1.0}
\setlength{\tabcolsep}{4pt}
\begin{tabular}{@{}lccccc|c@{}}
\toprule
\textbf{Domain} & \textbf{T1} & \textbf{T2} & \textbf{T3} & \textbf{T4} & \textbf{T5} & \textbf{Total} \\
\midrule
ogbn-arxiv      & 1{,}400 & 1{,}400 & 1{,}400 & 1{,}400 & 1{,}400 & 7{,}000 \\
ogbn-products   & 1{,}400 & 1{,}400 & 1{,}400 & 1{,}400 & 1{,}400 & 7{,}000 \\
PubMed          & 1{,}400 & 1{,}400 & 1{,}400 & 1{,}400 & 1{,}400 & 7{,}000 \\
WikiCS          & 1{,}400 & 1{,}400 & 1{,}400 & 1{,}400 & 1{,}400 & 7{,}000 \\
USPTO           & 1{,}400 & 1{,}400 & 1{,}400 & 1{,}400 & 1{,}400 & 7{,}000 \\
Physics SE      & 1{,}400 & 1{,}400 & 1{,}400 & 1{,}400 & 1{,}400 & 7{,}000 \\
\midrule
\textbf{Total}  & 8{,}400 & 8{,}400 & 8{,}400 & 8{,}400 & 8{,}400 & \textbf{42{,}000} \\
\bottomrule
\end{tabular}
\end{table}

\section{Experiments}
\label{sec:exp}

\noindent Table~\ref{tab:findings_summary} summarises the key findings
of this section up front. Each is supported by a paragraph in the
relevant subsection.

\begin{table}[h]
\centering
\caption{Key findings of \S\ref{sec:exp}, one per subsection.}
\label{tab:findings_summary}
\scriptsize
\setlength{\tabcolsep}{4pt}
\renewcommand{\arraystretch}{1.0}
\begin{tabular}{@{}llp{10.5cm}@{}}
\toprule
\textbf{Where} & \textbf{Topic} & \textbf{Headline finding} \\
\midrule
\S\ref{sec:main_results}      & Main Results          & \textbf{No family closes the gap.} LLMs compete on Description; only frontier LLMs clear the GNN on outlier detection; community detection stays with the GNN. \\
\S\ref{sec:template_ablation} & Template ablation     & \textbf{Denser context unlocks Description and outlier detection.} 1-hop neighbour titles unlock Description, 2-hop unlocks outlier detection, community detection stays stubborn. \\
\S\ref{sec:abl_reasoning}     & Reasoning supervision & \textbf{Reasoning carries the alignment signal.} Stripping reasoning collapses graph-token methods and lifts Graph2Text SFT. \\
\midrule
App.~\ref{sec:sft_scaling}    & SFT scaling to 70B    & \textbf{Capacity is not the bottleneck.} Output discipline is. \\
App.~\ref{sec:abl_pretraining}& Projector pretraining & \textbf{Pretraining helps wide label spaces.} Effect concentrates on Description with many fine-grained labels. \\
App.~\ref{sec:abl_gnn_backbone}& GNN encoder choice   & \textbf{Encoder matters where labels are wide.} The operator's inductive bias is load-bearing only with many fine-grained labels. \\
\bottomrule
\end{tabular}
\renewcommand{\arraystretch}{1.0}
\end{table}

\subsection{Evaluated Models}
\label{sec:models}

We compare four families of baselines on \bench. Hyperparameters,
training schedules, prompt templates, and hardware are deferred to
App.~\ref{app:impl}.

\textbf{Graph-token alignment.} Seven alignment models that
present the graph to the LLM as soft tokens or injected hidden
states: LLaGA~\cite{chen2024llaga},
GraphToken~\cite{perozzi2024let},
GraphGPT~\cite{tang2024graphgpt}, TEA-GLM~\cite{wang2024llms},
RGLM~\cite{zhang2026toward}, GOFA~\cite{kong2024gofa}, and
InstructGLM~\cite{ye2024language}. We hold the node-text encoder,
the GNN message-passing operator, and the LLM backbone fixed
across these baselines so any cross-baseline difference comes
from the loss, the projector, or the schedule.

\textbf{Zero-shot frontier LLM.}
GPT-5~\cite{openai2025gpt5} and Claude
Opus~4.7~\cite{anthropic2026opus47}. The subgraph is serialised as
plain text (an indexed node list and an edge list) and the prompt
provides only the canonical answer-frame sentence with no
closed-set label list.

\textbf{Graph2Text SFT.} Three open-weight LLMs QLoRA-fine-tuned
on the per-domain training split with the reference
Answer-plus-Reasoning target: Vicuna-7B, Qwen2-7B, and Llama-3-8B.

\textbf{Plain GNN.} GCN, GAT, and GraphSAGE trained on the same
training split. These isolate how much structural signal is
recoverable from graph topology alone under the same data budget.
Any graph-LLM method that fails to clear this reference cannot
claim its design adds anything over a purely structural solution.

\subsection{Evaluation Metrics}
\label{sec:metrics}

We use two complementary classes of metrics, both computed
deterministically from the structured per-line output with no LLM
judge in the loop. \textbf{Hard metrics} parse the answer block
and compare it against the structural ground truth: label accuracy
on T1, T2, and T3, per-node Hit@1 on T4, and ARI and NMI on T5.
\textbf{SBERT cosine} on the predicted versus gold \emph{Reasoning}
paragraph captures the soft side. We present
\texttt{all-mpnet-base-v2} (768d) in the main table because it
gives the largest spread between baselines. The
\texttt{all-MiniLM-L6-v2} and \texttt{e5-large-v2} numbers and
their ranking-stability check are in the appendix. We score the
Reasoning paragraph rather than the full Answer because the
Reasoning is the discriminative part. The two views are
complementary. A model can produce plausible cluster text
(moderate SBERT) yet miss the structural partition (low NMI),
so we report hard and soft side by side throughout.

\subsection{Main Results}
\label{sec:main_results}

\begin{table*}[t]
\centering
\caption{Main results, averaged across six domains. n{=}300 test
samples per (domain, task) cell.
SBERT columns report \texttt{all-mpnet-base-v2} cosine on the
predicted Reasoning paragraph. The plain-GNN rows show \texttt{n/a}
in every SBERT column because GNNs emit a label or a cluster
assignment, not a Reasoning paragraph to score against.
\textbf{Per-domain and per-encoder breakdowns are in the
appendix.}}
\label{tab:main}
\footnotesize
\renewcommand{\arraystretch}{1.05}
\setlength{\tabcolsep}{2.8pt}
\begin{tabular}{@{}l cc cc cc cc cc@{}}
\toprule
& \multicolumn{6}{c}{\textbf{Description}}
  & \multicolumn{4}{c}{\textbf{Comparison (harder)}} \\
\cmidrule(lr){2-7}\cmidrule(lr){8-11}
& \multicolumn{2}{c}{\textbf{T1}} & \multicolumn{2}{c}{\textbf{T2}}
  & \multicolumn{2}{c}{\textbf{T3}} & \multicolumn{2}{c}{\textbf{T4}}
  & \multicolumn{2}{c}{\textbf{T5}} \\
\cmidrule(lr){2-3}\cmidrule(lr){4-5}\cmidrule(lr){6-7}\cmidrule(lr){8-9}\cmidrule(lr){10-11}
\textbf{Model}
  & Acc. & SBERT
  & Acc. & SBERT
  & Acc. & SBERT
  & H@1 & SBERT
  & NMI & SBERT \\
\midrule
\multicolumn{11}{l}{\textit{Graph-token: GNN-LLM Aligned}} \\
\midrule
  LLaGA          & 0.218 & 0.558 & 0.654 & 0.554 & 0.276 & 0.549 & 0.066 & 0.346 & \textbf{0.066} & 0.443 \\
  GraphToken     & 0.164 & 0.533 & \underline{0.766} & 0.537 & 0.205 & 0.494 & \underline{0.101} & 0.260 & 0.047 & 0.394 \\
  GraphGPT       & 0.115 & 0.503 & 0.707 & 0.555 & 0.125 & 0.451 & 0.094 & 0.368 & \underline{0.061} & 0.441 \\
  TEA-GLM        & 0.199 & 0.549 & 0.721 & \underline{0.561} & 0.248 & 0.508 & 0.099 & \textbf{0.464} & 0.054 & \textbf{0.538} \\
  RGLM           & 0.219 & 0.561 & \textbf{0.783} & 0.560 & 0.282 & 0.526 & 0.097 & 0.329 & 0.050 & 0.465 \\
  GOFA           & \underline{0.339} & \underline{0.601} & \textbf{0.783} & \textbf{0.621} & \underline{0.441} & \underline{0.587} & \textbf{0.102} & \underline{0.396} & 0.042 & \underline{0.481} \\
  InstructGLM    & \textbf{0.539} & \textbf{0.618} & 0.495 & 0.463 & \textbf{0.800} & \textbf{0.596} & 0.080 & 0.324 & 0.026 & 0.182 \\
\midrule
\multicolumn{11}{l}{\textit{Closed-source frontier LLM (zero-shot)}} \\
\midrule
  GPT-5 (zero-shot)            & \underline{0.433} & \underline{0.762} & \underline{0.783} & \underline{0.743} & \textbf{0.700} & \underline{0.798} & \underline{0.458} & \textbf{0.705} & \underline{0.260} & \textbf{0.844} \\
  Claude Opus 4.7 (zero-shot)  & \textbf{0.467} & \textbf{0.774} & \textbf{0.992} & \textbf{0.754} & \textbf{0.700} & \textbf{0.804} & \textbf{0.642} & \underline{0.692} & \textbf{0.311} & \underline{0.748} \\
\midrule
\multicolumn{11}{l}{\textit{Graph2Text: SFT}} \\
\midrule
  SFT (Vicuna-7B)        & \textbf{0.580} & 0.808 & 0.747 & 0.811 & \underline{0.753} & 0.848 & 0.219 & 0.640 & 0.130 & \textbf{0.558} \\
  SFT (Qwen2-7B)         & 0.481 & \underline{0.833} & \underline{0.788} & \underline{0.821} & 0.698 & \textbf{0.864} & \underline{0.255} & \underline{0.662} & \underline{0.221} & 0.459 \\
  SFT (Llama-3-8B)       & \underline{0.550} & \textbf{0.846} & \textbf{0.821} & \textbf{0.844} & \textbf{0.766} & \underline{0.863} & \textbf{0.390} & \textbf{0.694} & \textbf{0.276} & \underline{0.477} \\
\midrule
\multicolumn{11}{l}{\textit{GNN: Structural reference}} \\
\midrule
  GCN              & \underline{0.608} & n/a & \textbf{0.849} & n/a & \textbf{0.892} & n/a & 0.117 & n/a & 0.239 & n/a \\
  GAT              & \textbf{0.613} & n/a & 0.588 & n/a & \underline{0.889} & n/a & \underline{0.228} & n/a & \underline{0.268} & n/a \\
  GraphSAGE        & 0.605 & n/a & \underline{0.723} & n/a & 0.885 & n/a & \textbf{0.412} & n/a & \textbf{0.457} & n/a \\
\bottomrule\end{tabular}
\renewcommand{\arraystretch}{1.0}
\end{table*}

Table~\ref{tab:main} reports each family's behaviour across the
five tasks. We walk through the four families in turn, then close
with what the joint pattern says about graph inference.

\paragraph{Graph-token alignment.} \textbf{Aligned LLMs can
describe a region of the graph but cannot compare regions.}
InstructGLM is the family's strongest model on Description (T1
$0.539$, T3 $0.800$). The whole family lands under $0.11$ on T4
and under $0.07$ on T5, two orders of magnitude below GraphSAGE.
The encoder is fine. The bottleneck sits downstream of it.

\paragraph{Frontier closed-source LLMs (zero-shot).}
\textbf{Zero-shot frontier LLMs dominate single-node Comparison,
still behind the GNN on multi-node partitioning.} Claude
Opus~4.7 takes T4 $0.642$, ahead of every other family including
plain GraphSAGE ($0.412$), and reaches T2 $0.992$.
Open-vocabulary Description and T5 multi-node partitioning still
trail SFT and the GNN respectively.

\paragraph{Graph2Text SFT.} \textbf{Updating the LLM weights
closes the Description gap, not the multi-node Comparison gap.}
QLoRA SFT on Llama-3-8B reaches T3 $0.766$, within a few points
of the GCN ceiling, and SFT-Vicuna leads T1 across the whole
table. T4 lands between the graph-token floor and the Claude
ceiling. T5 still trails GraphSAGE.

\paragraph{Plain GNN (structural reference).} \textbf{The
structural signal lives in the subgraph and a 2-layer GNN
extracts it, with no notion of relation semantics or
text-grounded comparison.} GCN, GAT, and GraphSAGE win three of
the five tasks (T1, T3, T5). T2 and T4 reward the language prior
more than the topology, and the GNN trails Claude there.

\paragraph{Joint takeaway.} Graph inference is not one
capability. On Description, LLM-based methods compete with the
GNN. On single-node Comparison, the frontier LLM is the only
method that clears the GNN. On multi-node Comparison, every
LLM-based method plateaus around 0.3 NMI while a small GNN
reaches 0.46. \emph{The signal is in the graph. Closing the
remaining gap is an objective and decoding problem, not a
capacity problem.}

\subsection{Better graph token template to fill the gap}
\label{sec:template_ablation}

\noindent \textbf{Adding explicit per-node neighbour titles to
the prompt recovers the
description-task signal that graph tokens alone fail to deliver,
and 2-hop context is what specifically unlocks the single-node
Comparison task.} We augment the LLaGA and GOFA prompts with
\textbf{h1} (1-hop neighbour titles) and \textbf{h2}
(1- and 2-hop neighbour titles), keeping the projector and the
Llama-3-8B backbone identical to the main-table rows. Both
variants are LoRA-fine-tuned on the same training split (shown below).

\begin{promptblock}{Template with interleaved graph token and titles}
\scriptsize
\textbf{LLaGA (graph tokens only):}
\begin{verbatim}
Node 0 <gt_0>      ...      Node N-1 <gt_{N-1}>      <task question>
\end{verbatim}
\textbf{h1 (GT + 1-hop titles):} each 1-hop neighbour's graph
token is followed by its title on the same line:
\begin{verbatim}
Node 0 <gt_0>: <title 0>
Node 1 <gt_1>: <title 1>      (1-hop neighbour of node 0)
Node 2 <gt_2>: <title 2>      (1-hop neighbour of node 0)
...                           <task question>
\end{verbatim}
\textbf{h2 (GT + 1+2-hop titles):} same as h1, plus a ``Node $k$
\texttt{<gt}$_k$\texttt{>}: title'' line for every 2-hop
neighbour of node 0. Projector, Llama-3-8B backbone, and LoRA
adapters are identical across variants; only title placement
changes.
\end{promptblock}

\begin{table}[h]
\centering
\caption{Template ablation on LLaGA and GOFA. \textbf{h1} adds
1-hop neighbour titles to the prompt. \textbf{h2} adds 1- and
2-hop. \textbf{GraphSAGE} is the no-LLM floor. \textbf{SFT
(Llama-3-8B)} is the full-LLM ceiling. n{=}300 per task. Per-domain
breakdowns of the reference rows are in
Tab.~\ref{tab:domain_gnn} (GraphSAGE),
Tab.~\ref{tab:appendix_sft_t123} (SFT), and
Tab.~\ref{tab:appendix_expanded_t123} (LLaGA, GOFA).}
\label{tab:template_ablation}
\scriptsize
\renewcommand{\arraystretch}{1.0}
\setlength{\tabcolsep}{2.5pt}
\begin{tabular}{@{}llccccc@{}}
\toprule
Domain & Variant & T1 acc & T2 acc & T3 acc & T4 H@1 & T5 NMI \\
\midrule
\multirow{8}{*}{ogbn-arxiv}
  & GraphSAGE \emph{(GNN ref.)}     & 0.529 & 0.605 & 0.812 & 0.393 & \textbf{0.255} \\
  & Llama-3-8B \emph{(SFT ref.)}    & 0.467 & \textbf{0.854} & \textbf{0.713} & 0.456 & 0.185 \\
\cmidrule(l){2-7}
  & LLaGA                           & 0.477 & 0.727 & 0.740 & 0.104 & 0.029 \\
  & LLaGA + h1 (GT + 1-hop)         & 0.617 & 0.780 & 0.887 & 0.143 & 0.062 \\
  & LLaGA + h2 (GT + 1+2-hop)       & 0.650 & 0.829 & 0.930 & 0.563 & 0.102 \\
  & GOFA                            & 0.042 & 0.737 & 0.062 & 0.112 & 0.028 \\
  & GOFA + h1 (GT + 1-hop)          & 0.620 & 0.733 & 0.923 & 0.563 & 0.104 \\
  & GOFA + h2 (GT + 1+2-hop)        & \textbf{0.663} & 0.810 & \textbf{0.933} & \textbf{0.583} & 0.073 \\
\midrule
\multirow{8}{*}{patents}
  & GraphSAGE \emph{(GNN ref.)}     & 0.450 & 0.699 & 0.853 & 0.201 & \textbf{0.300} \\
  & Llama-3-8B \emph{(SFT ref.)}    & 0.366 & 0.857 & 0.593 & 0.273 & 0.254 \\
\cmidrule(l){2-7}
  & LLaGA                           & 0.063 & 0.763 & 0.047 & 0.067 & 0.082 \\
  & LLaGA + h1 (GT + 1-hop)         & 0.497 & \textbf{0.879} & 0.883 & 0.104 & 0.130 \\
  & LLaGA + h2 (GT + 1+2-hop)       & \textbf{0.503} & 0.874 & 0.887 & 0.341 & 0.167 \\
  & GOFA                            & 0.055 & 0.788 & 0.048 & 0.055 & 0.075 \\
  & GOFA + h1 (GT + 1-hop)          & 0.173 & 0.708 & 0.887 & 0.253 & 0.235 \\
  & GOFA + h2 (GT + 1+2-hop)        & 0.477 & 0.863 & \textbf{0.920} & \textbf{0.350} & 0.128 \\
\midrule
\multicolumn{7}{@{}l}{\emph{LLaGA-h2 generalization to additional domains (n{=}300/task):}} \\
ogbn-products    & h2 (GT + 1+2-hop)              & 0.743 & 0.827 & 0.887 & 0.385 & 0.116 \\
physics-se       & h2 (GT + 1+2-hop)              & 0.523 & 0.959 & 0.847 & 0.547 & 0.107 \\
pubmed-diabetes  & h2 (GT + 1+2-hop)              & 0.837 & 0.870 & 0.990 & 0.559 & 0.213 \\
wikics           & h2 (GT + 1+2-hop)              & 0.827 & 0.850 & 0.977 & 0.483 & 0.367 \\
\bottomrule
\end{tabular}
\end{table}

\paragraph{1-hop unlocks Description, 2-hop unlocks T4.} On
patents the GT-only LLaGA collapses to T1 $0.063$ and T3
$0.047$. Adding 1-hop neighbour titles lifts T1 to $0.497$ and
T3 to $0.883$. The graph token alone under-determines the
answer. One hop of explicit text fills the gap. Adding the
second hop adds little to T1 / T3 but lifts T4 specifically:
LLaGA T4 H@1 jumps from $0.10$ to $0.56$ on arxiv (+46 percentage points) and
from $0.07$ to $0.34$ on patents (+27 percentage points), and the GOFA rows
repeat the same pattern under a different alignment recipe. The
h1 setting barely moves T4 over GT-only, so the T4 gain is
specifically a 2-hop effect. Outlier detection requires
comparing each node against its neighbours' neighbours, and 1-hop
context cannot supply that.

\paragraph{T5 stays stubborn.} NMI rises monotonically from
LLaGA to h1 to h2 but stays well below the GNN reference on both
domains. Denser prompt context lifts every other task close to or
above the GNN ceiling, yet does not move multi-node partitioning,
which a 2-layer GNN extracts directly from the topology. T5
remains a structural-coverage problem that prompt augmentation
alone cannot solve.

\subsection{Does the Reasoning supervision help?}
\label{sec:abl_reasoning}

\noindent \textbf{Stripping the Reasoning paragraph from the
training target collapses graph-token methods on T1, T3, and T4
but does not hurt Graph2Text SFT, indicating that
supervision-side scaffolding is what gets the alignment recipe to
read the graph at all.} We retrain LLaGA, TEA-GLM, and GOFA on
ogbn-arxiv and pubmed-diabetes asking the model to emit only
\texttt{Answer: \textit{X}}, with everything else unchanged.

\begin{figure}[h]
\centering
\includegraphics[width=\linewidth]{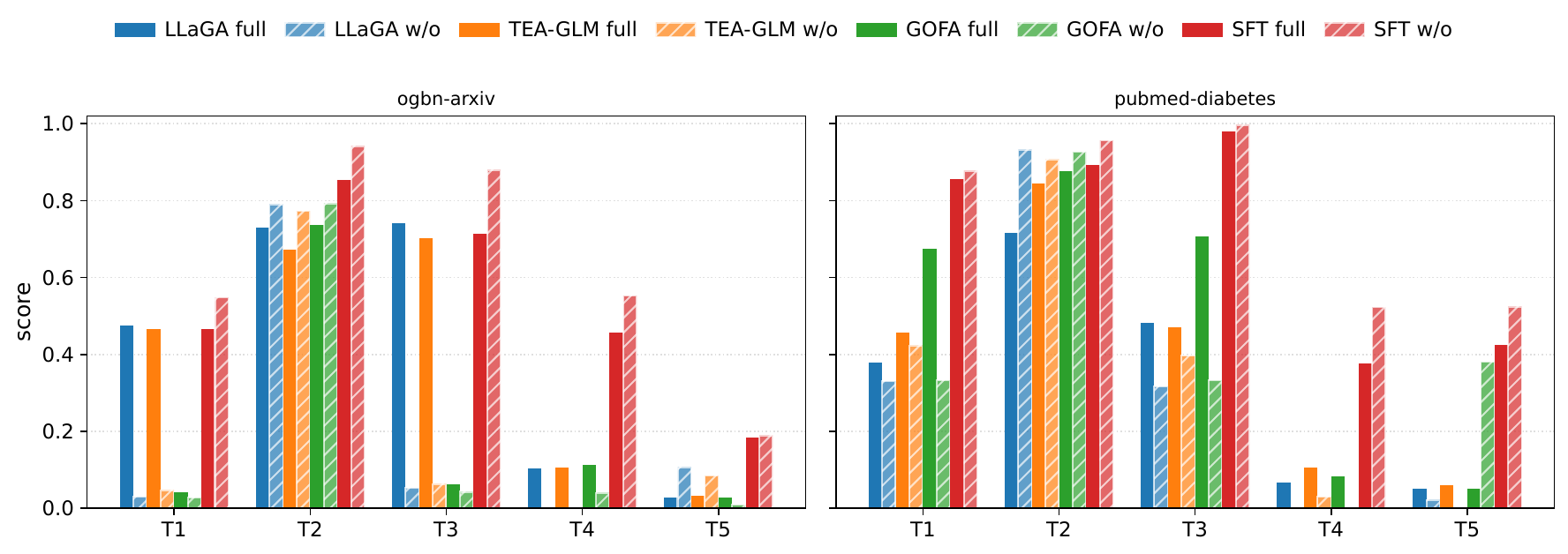}
\caption{Reasoning-supervision ablation. \textbf{full} is the
main-table recipe (Answer plus Reasoning). \textbf{w/o} strips
the Reasoning paragraph. n{=}300 per task. Per-domain numbers
for the \textbf{full} bars are in
Tab.~\ref{tab:appendix_expanded_t123} (LLaGA, TEA-GLM, GOFA) and
Tab.~\ref{tab:appendix_sft_t123} (SFT).}
\label{fig:abl_reasoning_bars}
\end{figure}

\paragraph{Reasoning carries the Description signal.} Hard
accuracy on T1 (masked node) and T3 (theme summarisation)
collapses by tens of percentage points on every graph-token
(model, domain) pair we test. Without the rationale paragraph,
the projector and LoRA jointly fit a shorter target distribution
that captures the answer phrase as a generic surface form. What
gets lost is the per-instance grounding to the graph, which
emerges only when the model is forced to articulate \emph{why}
the label is correct.

\paragraph{T4 collapses for graph-token, lifts for SFT.} For
graph-token baselines, removing Reasoning drives T4 H@1 to near
zero. Inspecting predictions, the model emits ``Node $k$: not
outlier'' for every node, so it never names any outlier.
Graph2Text SFT goes the other way: T4 \emph{improves} without
Reasoning. The full-LLM gradient of SFT lets the model
discriminate from graph-text alone, and removing the free-form
rationale removes a competing gradient. T2 (yes/no) also gains a
few percentage points without Reasoning across all rows for the
same reason. The ablation is therefore an honest signal, not a
uniform regression: it surfaces a real supervision-vs-decoding
asymmetry between the two paradigms.

\subsection{More findings}
\label{sec:findings_appendix}

Three additional
ablations are reported in the appendix and reinforce the same
narrative. \textbf{Projector pretraining} unlocks Description on
the wide label space and is a no-op on the narrow one
(App.~\ref{sec:abl_pretraining}). \textbf{Scaling SFT from 8B to
70B} helps constrained-output tasks and hurts open-vocabulary
description, indicating that capacity is not the bottleneck on
Description (App.~\ref{sec:sft_scaling}). \textbf{The GNN encoder
choice} flips with label-space size: SAGE wins on the 40-class
arxiv space, GCN slightly wins on 3-class pubmed
(App.~\ref{sec:abl_gnn_backbone}).

\section{Conclusion}

\bench\ measures the capability of LLMs to read a node and its
neighbours and produce an open-ended conclusion that no single node
implies, framed along two axes: \emph{Description} (verbalising what
a region is, T1 to T3) and \emph{Comparison} (distinguishing how
regions differ, T4 and T5). Five tasks over six structurally diverse
domains and a four-layer quality-control pipeline yield a public
release of \textbf{42{,}000 samples} with deterministic ground truth
at zero annotation cost, evaluated under four baseline families on
the same splits (graph-token aligned LLMs, zero-shot frontier LLMs,
Graph2Text SFT, and plain GNNs). \textbf{Frontier LLMs and
Graph2Text SFT close the description-task gap, but no LLM-based
method recovers the multi-node partition signal that plain GNNs
extract}, leaving \bench\ as the diagnostic tool to track whether
graph-LLM alignment and text-serialised LLM approaches can close
that remaining gap.

\smallskip\noindent
Code: \url{https://github.com/graphinfer/GraphInfer-Bench}.\hspace{0.6em}%
Dataset: \url{https://huggingface.co/datasets/graphinfer/graphinfer}.

\bibliographystyle{plainnat}
\bibliography{references_corrected}

\newpage
\appendix

\section{Limitations}
\label{app:limitations}

\bench currently spans six text-attributed graph domains: academic
citations, e-commerce co-purchase, biomedical citations, encyclopedia
articles, patent citations, and a Q\&A community. Two domains we
would have liked to include are absent: \textbf{finance} (e.g.,
transaction or trading networks) and \textbf{biology} beyond
biomedical citations (e.g., protein-protein interaction or single-cell
graphs with rich textual annotations). Both are difficult to obtain at
scale: financial transaction graphs are typically proprietary or
subject to strict redistribution constraints, and biology graphs with
the per-node free-text descriptions our pipeline requires are scarce
in the public domain. Adding these two families is the most direct
extension of the benchmark.

\section{Broader Impacts}
\label{app:broader_impacts}

\paragraph{Positive impacts.} {\bench} standardises evaluation for
graph-LLM reasoning under a unified split, prompt, and scoring
protocol, removing two confounds that have plagued the field
(non-comparable domain choices and per-paper metric variations).
This makes incremental progress easier to recognise and harder to
overstate. The dataset, the gold reasoning, and the audit
annotations are released under open licences (CC BY 4.0 for data,
MIT for code, App.~\ref{app:license}), so any group can reproduce
the leaderboard or extend it. By identifying the multi-node
partition gap (\S\ref{sec:exp}) and the description-vs-comparison
asymmetry, the benchmark gives downstream research concrete failure
modes to target rather than aggregate scores to chase.

\paragraph{Negative impacts.} The gold reasoning is generated by
DeepSeek for the candidate pool, with Llama-3.1-70B and
Qwen-2.5-72B used for Layer-2 audit. Any of these models may
carry biases or blind spots that propagate into the gold target. Layer 1 to Layer 3 of the
quality-control pipeline (\S\ref{sec:quality}) catch many surface
defects but do not certify substantive bias. Any user training on
\bench inherits those biases. A second concern is dual-use: a strong
graph-token model trained against this benchmark could be applied to
sensitive social or financial graphs in ways that we cannot foresee,
particularly if the model is deployed without further evaluation on
the target domain. Finally, training and judging large LLMs has a
non-trivial energy cost; we report compute resources
(App.~\ref{app:release_gate_l2}, App.~\ref{sec:sft_scaling}) so
practitioners can decide whether the benefit justifies the cost.

\section{Does projector pretraining help?}
\label{sec:abl_pretraining}

\noindent \textbf{Projector pretraining unlocks Description on
the larger label space, and is a no-op on the smaller one.} Three
of the seven graph-token baselines (TEA-GLM, GOFA, RGLM) include a
projector or encoder pretraining pass before joint fine-tuning. We
disable that step, keeping every other hyperparameter identical,
on ogbn-arxiv (40 classes) and pubmed-diabetes (3 classes).

\begin{figure}[h]
\centering
\includegraphics[width=\linewidth]{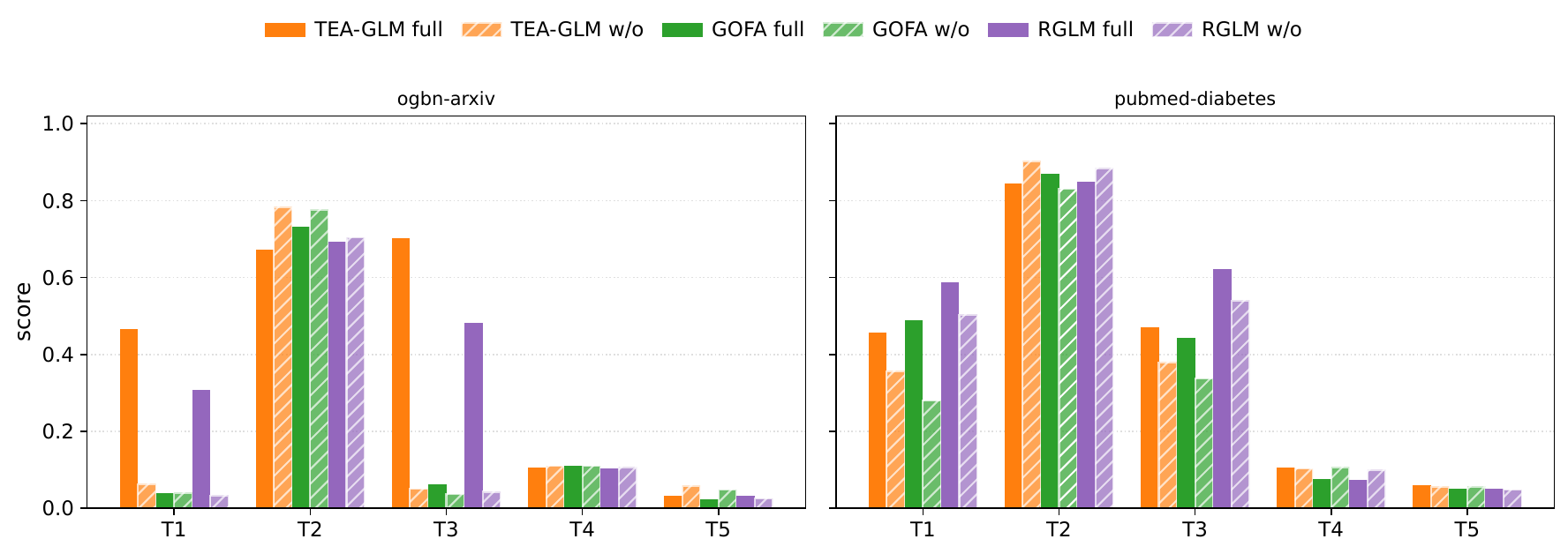}
\caption{Projector and encoder pretraining ablation on TEA-GLM,
GOFA, and RGLM. Solid bars are the \textbf{full} main-table
recipe; hatched bars skip the corresponding pretraining pass and
train directly from raw SBERT. n{=}300 per task, mean across 3
seeds. The drop concentrates on T1 and T3 (description tasks),
strongest on the 40-class arxiv space. Per-domain numbers for the
\textbf{full} rows are in Tab.~\ref{tab:appendix_expanded_t123}.}
\label{fig:abl_pretraining_bars}
\end{figure}

\paragraph{Effect concentrates on Description in the wide label
space.} Disabling TEA-GLM's PCA alignment collapses the 40-class
arxiv space (T1 $0.47$ to $0.06$, T3 $0.70$ to $0.05$) and gives a
smaller drop on the 3-class pubmed space (T1 $0.46$ to $0.36$, T3
$0.47$ to $0.38$). GOFA shows the same shape on pubmed (T1 $0.49$
to $0.28$, T3 $0.44$ to $0.34$). T2, T4, and T5 are invariant
across the ablation. Pretraining is doing lexical disambiguation
in the projector rather than adding structural capacity, and the
benefit scales with how many fine-grained labels the projector has
to separate. With \S\ref{sec:abl_reasoning}, the description-task
signal in graph-token methods comes from supervision-side
scaffolding (Reasoning paragraphs and projector pretraining), not
from raw graph-token capacity.

\section{Does scaling SFT to 70B help?}
\label{sec:sft_scaling}

\noindent \textbf{Scaling Graph2Text SFT from 8B to 70B helps
constrained-output tasks and hurts open-vocabulary description.}
We re-run the SFT recipe on Llama-3.1-70B-Instruct-AWQ
($\sim 10\times$ the parameters), keeping the same training
data, prompts, and test split, on the two hardest domains
(ogbn-arxiv and patents).

\begin{table}[h]
\centering
\caption{SFT scaling: 8B vs.\ 70B on the two hardest domains.
n{=}300 per task, mean$\pm$std across 3 seeds. Per-domain 8B
reference numbers are in Tab.~\ref{tab:appendix_sft_t123}.}
\label{tab:sft_scaling}
\small
\setlength{\tabcolsep}{3.5pt}
\begin{tabular}{@{}llccccc@{}}
\toprule
Domain & Model & T1 acc & T2 acc & T3 acc & T4 H@1 & T5 NMI \\
\midrule
\multirow{2}{*}{ogbn-arxiv}
  & SFT (Llama-3-8B)  & \textbf{0.467\,$\pm$\,0.015} & 0.854\,$\pm$\,0.025 & \textbf{0.713\,$\pm$\,0.007} & 0.456\,$\pm$\,0.013 & \textbf{0.185\,$\pm$\,0.006} \\
  & SFT (Llama-3-70B)    & 0.319\,$\pm$\,0.027          & \textbf{0.909\,$\pm$\,0.016} & 0.493\,$\pm$\,0.027          & \textbf{0.503\,$\pm$\,0.030} & 0.167\,$\pm$\,0.011 \\
\midrule
\multirow{2}{*}{patents}
  & SFT (Llama-3-8B)  & 0.366\,$\pm$\,0.027          & 0.857\,$\pm$\,0.012 & \textbf{0.593\,$\pm$\,0.028} & 0.273\,$\pm$\,0.001 & \textbf{0.254\,$\pm$\,0.022} \\
  & SFT (Llama-3-70B)    & \textbf{0.520\,$\pm$\,0.029} & \textbf{0.913\,$\pm$\,0.016} & 0.467\,$\pm$\,0.029          & \textbf{0.275\,$\pm$\,0.025} & 0.235\,$\pm$\,0.011 \\
\bottomrule
\end{tabular}
\end{table}

\paragraph{70B helps where the output is constrained.} T2 lifts
on both domains (arxiv $0.85$ to $0.91$, patents $0.86$ to $0.91$).
T4 lifts on arxiv ($0.46$ to $0.50$). Both have constrained
outputs (yes/no, a single node ID) and reward the larger
language prior.

\paragraph{70B hurts on open-vocabulary description.} T1 and T3
drop on most cells (arxiv T1 $0.47$ to $0.32$, arxiv T3
$0.71$ to $0.49$, patents T3 $0.59$ to $0.47$). SBERT-Reason
cosines sit within $0.05$ of the 8B row everywhere, so the
graph-reading quality is unchanged. The accuracy loss is on the
output side: AWQ-INT4 quantisation widens the output
distribution and the 70B language prior paraphrases the
canonical label, both of which hurt tightly-templated
description targets but leave Yes/No or single node ID intact.
Capacity is not the bottleneck on Description. Output discipline
is.

\section{Does the GNN encoder choice matter?}
\label{sec:abl_gnn_backbone}

\noindent \textbf{The message-passing operator is load-bearing
only when the projector has many fine-grained labels to
disambiguate, mirroring the pretraining ablation in the main
paper.} The main table fixes GraphSAGE as the unified GNN choice
across GraphToken, TEA-GLM, RGLM, and GOFA so that any
cross-baseline differences live in the loss, pretrain, or
projection head rather than the operator. Here we re-train
GraphToken on ogbn-arxiv and pubmed-diabetes, swapping
\texttt{SAGEConv} for \texttt{GCNConv} with everything else fixed.

\begin{table}[h]
\centering
\caption{GNN-backbone ablation on GraphToken. \textbf{SAGE} is the
canonical main-table operator. \textbf{GCN} is the ablation.
n{=}300 per task, mean$\pm$std across 3 seeds. Per-domain
SBERT-Reason cosines for the \textbf{SAGE} reference are in
Tab.~\ref{tab:appendix_expanded_t123}.}
\label{tab:abl_gnn_backbone}
\small
\setlength{\tabcolsep}{4pt}
\begin{tabular}{@{}llccccc@{}}
\toprule
Domain & GNN & T1 acc & T2 acc & T3 acc & T4 H@1 & T5 NMI \\
\midrule
\multirow{2}{*}{ogbn-arxiv}
  & SAGE & \textbf{0.383\,$\pm$\,.028} & 0.670\,$\pm$\,.027 & \textbf{0.677\,$\pm$\,.027} & \textbf{0.113\,$\pm$\,.018} & \textbf{0.038} \\
  & GCN  & 0.050\,$\pm$\,.013          & \textbf{0.683\,$\pm$\,.027} & 0.043\,$\pm$\,.012 & 0.106\,$\pm$\,.018 & 0.034 \\
\midrule
\multirow{2}{*}{pubmed-diabetes}
  & SAGE & 0.587\,$\pm$\,.028          & 0.850\,$\pm$\,.021 & 0.617\,$\pm$\,.028 & \textbf{0.087\,$\pm$\,.016} & \textbf{0.055} \\
  & GCN  & \textbf{0.667\,$\pm$\,.027} & \textbf{0.886\,$\pm$\,.018} & \textbf{0.653\,$\pm$\,.027} & 0.084\,$\pm$\,.016 & 0.052 \\
\bottomrule
\end{tabular}
\end{table}

\paragraph{Encoder choice flips with the label space.} Swapping
SAGE to GCN on the 40-class arxiv space collapses GraphToken's T1
by $33$ percentage points ($0.38$ to $0.05$) and T3 by $63$
percentage points ($0.68$ to $0.04$), with parallel SBERT-Reason
drops (T1 $0.63$ to $0.32$, T3 $0.64$ to $0.27$). On 3-class
pubmed-diabetes the picture inverts: GCN slightly outperforms SAGE
on T1 ($0.67$ vs.\ $0.59$), T2 ($0.89$ vs.\ $0.85$), and T3
($0.65$ vs.\ $0.62$). The operator's inductive bias matters where
the projector has many fine-grained labels to separate, the same
pattern as pretraining (App.~\ref{sec:abl_pretraining}).
Cross-baseline parity benefits from fixing the operator as the
main table does, but the direction of any swap is
domain-dependent.

\section{License and Data Use}
\label{app:license}

\bench releases its derived data (ego-graph samples, gold labels,
gold reasoning, audit annotations) under \textbf{CC BY 4.0} and the
accompanying code under \textbf{MIT}. Both permit commercial use
with citation. Source graphs retain their upstream licences,
listed below. Users must respect these when accessing raw graph
structure beyond what we redistribute.

\begin{itemize}
  \item \textbf{ogbn-arxiv}: OGB code MIT; arXiv \textbf{title
        metadata} is CC0 1.0 Public Domain (the only field we use).
  \item \textbf{ogbn-products}: OGB code MIT; underlying Amazon
        co-purchase data from Bhatia et al.\ 2016 (Chiang et al.\
        2019), distributed for academic research use.
  \item \textbf{PubMed-Diabetes}: graph dataset (Sen et al.\ 2008)
        in academic public release; abstract text is not held under
        copyright by NLM but may be subject to publisher copyrights.
  \item \textbf{WikiCS}: dataset repository MIT (Mernyei \&
        Cangea 2020); underlying Wikipedia article text CC BY-SA
        4.0 (CC BY-SA 3.0 for contributions prior to June 2023).
  \item \textbf{USPTO patents}: not subject to U.S.\ copyright per
        17 USC \S 105; USPTO reserves the right to assert copyright
        internationally; bulk-data redistribution permitted via
        USPTO Open Data Portal.
  \item \textbf{Physics StackExchange}: CC BY-SA 4.0 (per Stack
        Exchange Network terms).
\end{itemize}

Gold reasoning and Layer-2 verdicts were produced by DeepSeek,
Llama-3.1-70B-Instruct (AWQ), and Qwen-2.5-72B-Instruct (AWQ).
Each provider permits redistribution as academic content, and we
manually inspected samples for harmful content during the L1 to L3
release gate (\S\ref{app:verification}). The benchmark contains no
PII beyond the upstream public corpora.

\section{Datasheet for GraphInfer-Bench}
\label{app:datasheet}

We follow the datasheet template of Gebru et al.\
(\href{https://arxiv.org/abs/1803.09010}{arXiv:1803.09010}).
Where the answer already lives elsewhere in the paper, we
cross-reference rather than duplicate.

\subsection{Motivation}

\paragraph{For what purpose was the dataset created?}
To evaluate the capability of an LLM to read a node and its
neighbourhood and produce an open-ended natural-language
conclusion that no single node implies (\S\ref{sec:intro}).
Existing graph-QA protocols (algorithm simulation, node
classification, single-node description, KG-QA, GraphRAG) cannot
test this capability (Tab.~\ref{tab:eval_gap}).

\paragraph{Who created the dataset and on behalf of which entity?}
The authors.

\paragraph{Who funded the creation of the dataset?}
N/A.

\subsection{Composition}

\paragraph{What do the instances represent?}
Each instance is a 2-hop ego-graph from one of six real
text-attributed graphs (Tab.~\ref{tab:domains}), paired with one
of five graph-inference tasks (Tab.~\ref{tab:tasks}) and a
reference Answer plus Reasoning ground truth.

\paragraph{How many instances are there in total?}
$42{,}000$ released samples: $6$ domains $\times$ $5$ tasks
$\times$ $1{,}400$ samples per cell. See Tab.~\ref{tab:stats}.

\paragraph{Does the dataset contain all possible instances or is it
a sample?}
A sample. The pre-quality-control candidate pool contains
$66{,}000$ candidates (App.~\ref{app:candidate_pool}); $42{,}000$
ship after quality control and per-cell capping.

\paragraph{What data does each instance consist of?}
An (input, answer, reasoning) triple. Inputs include the
ego-graph node list, edge list, per-node titles, and the task
question; the answer is the deterministic gold target; the
reasoning is a short rationale grounding the answer in specific
nodes and edges (\S\ref{sec:tasks}).

\paragraph{Is there a label or target associated with each
instance?}
Yes. T1, T3 emit a category label, T2 a yes/no, T4 a per-node
outlier label, T5 a partition assignment.

\paragraph{Is any information missing from individual instances?}
Yes, by design. For T1 the hub node's title is replaced with
\masked. Node abstracts and the dataset-level class label of
each node are deliberately withheld from model inputs so that the
shortcut of looking up a single node's metadata cannot solve the
task. See \S\ref{sec:tasks} and the per-task prompt templates in
App.~\ref{app:tasks}.

\paragraph{Are there recommended train/val/test splits?}
Yes. Each (domain, task) cell holds $1{,}000$ train, $100$ val,
and $300$ test (Tab.~\ref{tab:stats}).

\paragraph{Are there any errors, sources of noise, or redundancies?}
The Layer-1 to Layer-4 release gate admits $90.4\%$ of the
candidate pool (\S\ref{sec:quality}, App.~\ref{app:verification}).
Residual noise comes from the gold reasoning, which is generated
by an LLM and audited for surface defects but not certified for
substantive bias.

\paragraph{Does the dataset relate to people?}
No. The source graphs are public corpora (academic citations,
e-commerce co-purchase, biomedical citations, encyclopedia,
patents, physics Q\&A). The release contains no PII beyond what
is already public in those corpora.

\subsection{Collection Process}

\paragraph{How was the data acquired?}
The six source graphs are public datasets retrieved from their
upstream repositories under their respective licences
(App.~\ref{app:license}, App.~\ref{app:domains}).

\paragraph{What mechanisms or procedures were used to collect the
data?}
Per domain we sample 2-hop ego-graphs centred on hub nodes with
in-degree $\geq 3$ and valid node text, capping neighbours at
ten per hop (App.~\ref{app:candidate_pool}). Each ego-graph is
paired with one of the five tasks under deterministic per-task
construction rules (App.~\ref{app:tasks}).

\paragraph{If sampling, what was the sampling strategy?}
Random sampling without replacement among eligible hub nodes per
domain. We oversample to $2{,}200$ candidates per (domain, task)
cell so that the cell still meets its $1{,}400$-sample quota
after the quality-control pipeline rejects roughly $10\%$.

\paragraph{Over what timeframe was the data collected?}
Sampling, gold-reasoning generation, and the four-layer audit
were run between March and May 2026.

\paragraph{Was any data excluded? Why?}
Yes. An earlier seventh candidate domain, \texttt{string-db},
was dropped because its protein-interaction node descriptions
exceeded the prompt budget and produced a Layer-2 admission rate
below $14\%$ in pilot runs, which would not meet the per-cell
quota.

\paragraph{Were any ethical review processes conducted?}
N/A. No human-subject data are collected. The benchmark is built
on already-public corpora.

\subsection{Preprocessing, Cleaning, Labeling}

\paragraph{Was any preprocessing or cleaning done?}
Yes. A four-layer release gate (\S\ref{sec:quality},
App.~\ref{app:verification}) audits the gold reasoning under a
shared C1--C5 vocabulary (Layers 1--3) and deduplicates the
observed graph against split leakage (Layer 4).

\paragraph{Who generated the gold reasoning?}
DeepSeek (the API as of 2026-04-01,
\S\ref{sec:tasks} footnote 2). Layer-2 audit verdicts used to
filter the gold reasoning are produced by Llama-3.1-70B-Instruct
(AWQ) and Qwen-2.5-72B-Instruct (AWQ); see
App.~\ref{app:release_gate_l2}.

\paragraph{Who performed the Layer-3 human calibration?}
Two annotators drawn from the author team. No external
annotators were recruited and no compensation was paid
(App.~\ref{app:release_gate_l3}).

\paragraph{How were the SBERT acceptance thresholds calibrated?}
The per-domain thresholds in App.~\ref{app:tasks} (papers
$0.30$, products $0.25$, articles $0.25$, patents $0.25$,
questions $0.20$) were tuned on the validation (dev) split. The
test split was not used in threshold selection.

\paragraph{Was the raw data saved alongside the preprocessed
data?}
Yes. The release ships the $66{,}000$-sample candidate pool, the
Layer-1 / Layer-2 / Layer-3 verdicts, and the $42{,}000$-sample
released benchmark.

\subsection{Uses}

\paragraph{Has the dataset been used for any tasks already?}
Yes. \S\ref{sec:exp} reports four baseline families
(graph-token alignment, zero-shot frontier LLMs, Graph2Text SFT,
plain GNNs) under matched splits and a unified scoring protocol.

\paragraph{Is there a repository linking to papers or systems
that use the dataset?}
The code repository at
\url{https://github.com/graphinfer/GraphInfer-Bench} will
maintain a pointer to follow-on papers and leaderboards.

\paragraph{What (other) tasks could the dataset be used for?}
Diagnostics for retrieval-augmented LLM systems that ingest
graph context, ablation testbeds for new graph-token alignment
objectives, probes for emergent multi-node reasoning in larger
LLMs.

\paragraph{Are there tasks for which the dataset should NOT be
used?}
The dataset should not be used to train or fine-tune models for
direct deployment in finance or healthcare decision-making: the
gold reasoning is LLM-generated and does not constitute
financial or medical advice. It should not be treated as a
knowledge-graph QA benchmark; the questions probe inference, not
retrieval. The gold reasoning paragraph itself should not be
treated as ground-truth fact about the underlying domain.

\subsection{Distribution}

\paragraph{Will the dataset be distributed to third parties?}
Yes. \bench is released as a public benchmark.

\paragraph{How will the dataset be distributed?}
Dataset on Hugging Face Hub
(\url{https://huggingface.co/datasets/graphinfer/graphinfer});
code on GitHub
(\url{https://github.com/graphinfer/GraphInfer-Bench}).

\paragraph{When will the dataset be distributed?}
On paper acceptance.

\paragraph{Under what license?}
Derived data (samples, gold labels, gold reasoning, audit
annotations) under \textbf{CC BY 4.0}; code under \textbf{MIT}
(App.~\ref{app:license}).

\paragraph{Are there third-party IP restrictions on the data?}
Source graphs retain their upstream licences (ODC-BY 1.0 for
OGB, US public domain for PubMed-Diabetes and USPTO,
CC BY-SA 3.0 for WikiCS, CC BY-SA 4.0 for Physics
StackExchange). Users must respect these when accessing raw
graph structure beyond what we redistribute
(App.~\ref{app:license}).

\paragraph{Do export controls or regulatory restrictions apply?}
No.

\subsection{Maintenance}

\paragraph{Who will be supporting, hosting, maintaining the
dataset?}
The authors, on a best-effort basis.

\paragraph{How can the curator be contacted?}
GitHub issues at the code repository and dataset discussions on
the Hugging Face dataset page.

\paragraph{Is there an erratum?}
A \texttt{CHANGELOG.md} in the GitHub repository will track
post-release errata and corrections.

\paragraph{Will the dataset be updated?}
Yes. A v2 release is planned that adds finance and biology
domains (App.~\ref{app:limitations}). Old versions will be
tagged on Hugging Face and remain accessible. Updates will be
announced in the repository \texttt{CHANGELOG.md} and on the
Hugging Face dataset card.

\paragraph{If others want to extend, augment, or contribute, is
there a mechanism?}
Yes. External contributors can propose new domains, task
variants, or stronger gold-reasoning audits via GitHub pull
requests. Each proposal is reviewed under the same Layer-1 to
Layer-3 quality control as the original release.

\section{Implementation Details}
\label{app:impl}

This section collects training and inference details for every
baseline. Code and per-baseline launchers are released alongside
the dataset.

\paragraph{Common setup.} Node-text input features are SBERT
embeddings from \texttt{sentence-transformers/all-MiniLM-L6-v2}
(384d), shared across every alignment baseline so cross-baseline
differences are not confounded by node-text encoder choice. The
LLM backbone is \texttt{Llama-3-8B-Instruct} for the graph-token
family and the SFT family unless the original paper specified a
different model. Per-domain training splits are 1{,}000 samples
per task (5{,}000 total per domain), val 100 per task, test 300
per task.

\paragraph{Graph-token alignment baselines.} Stage 1 trains a
2-layer \texttt{SAGEConv} encoder with hidden 768. Stage 2 jointly
trains the projector and a LoRA on the LLM. We use the
``rounds'' schedule from the original LLaGA recipe: three rounds
of (one epoch projector-only, one epoch projector+LoRA), with
per-round evaluation and best-round-wins. LoRA rank 32 on
\texttt{q,k,v,o}. Optimizer AdamW, learning rate $5\times10^{-4}$,
linear warm-up 3\%, batch size 1 with gradient accumulation 8,
max sequence length 2{,}048. Eval uses vLLM with greedy decoding
and \texttt{max\_new\_tokens=384} (T1 to T4) or 2{,}500 (T5).

\paragraph{Zero-shot frontier LLMs.} GPT-5 and Claude Opus 4.7.
Subgraphs are serialised as a
plain-text indexed node list plus an edge list. The prompt
provides the canonical answer-frame sentence with no closed-set
label list. For T1 and T3 the rescorer projects the free-form
answer onto the per-domain canonical label set with
SBERT-MiniLM-L6 cosine before scoring. For Opus 4.7 we omit
\texttt{temperature} (the model uses its default); for GPT-5 we
set \texttt{reasoning\_effort=minimal} and pass
\texttt{max\_completion\_tokens=2048}.

\paragraph{Graph2Text SFT.} QLoRA (NF4, double-quant) on
Vicuna-7B-v1.5-16k, Qwen2-7B-Instruct, and Llama-3-8B-Instruct.
Training set is the 5{,}000 samples per domain rendered into
ShareGPT format with the answer-plus-reasoning target. LoRA rank
16, alpha 32, dropout 0.05, lr $2\times10^{-4}$, batch size 1
with gradient accumulation 8, 2 epochs, warm-up ratio 0.03. Eval
uses vLLM with greedy decoding and
\texttt{max\_new\_tokens=384}.

\paragraph{Plain GNN reference.} 2-layer GCN, GAT, GraphSAGE with
hidden 768, dropout 0.5, optimizer AdamW with lr $1\times10^{-3}$,
weight decay $5\times10^{-4}$, early stopping (patience 20) on
val accuracy. Three seeds (42, 43, 44). For GCN and GAT we use a
\texttt{concat\_input=True} variant that concatenates the raw
SBERT input to the final-layer GNN output before the head, which
gave the best results among the variants we tested.

\paragraph{Hardware.} Graph-token training and SFT-8B run on
single A100 (40\,GB) or A800 (80\,GB) GPUs. SFT-70B uses
Llama-3.1-70B-Instruct-AWQ-INT4 on a single RTX PRO 6000
Blackwell (97\,GB). Layer-2 dual judges run two AWQ-INT4 70B
models in parallel on the same Blackwell card pool. Per-baseline
training time is ${<}3$\,hours per (domain, task) cell. The full
benchmark and ablation suite consumed approximately 2{,}250 A100
GPU-hours.

\section{Per-Domain Raw Data}
\label{app:domains}

\paragraph{ogbn-arxiv (academic citation).}
169K CS papers with 40 subject-area labels~\cite{hu2020open}.
Directed citation edges. Rich abstracts make SBERT verification
straightforward; this is the primary development domain.
\textit{License.} OGB graph + node features:
\textbf{ODC-BY 1.0}; arXiv abstracts: per-paper, predominantly
\textbf{CC-BY 4.0} (and a smaller share of arXiv non-exclusive
licenses).

\paragraph{ogbn-products (e-commerce co-purchase).}
2.4M Amazon products with 47 category labels~\cite{hu2020open}.
Co-purchase edges encode consumer behaviour, not topical similarity.
Product titles are shorter than paper abstracts, so SBERT thresholds
are relaxed $\sim$20\%.
\textit{License.} OGB graph + node features:
\textbf{Amazon-Berkeley Objects custom research license} (research
use permitted; redistribution allowed within OGB).

\paragraph{PubMed (clinical citation).}
19{,}717 papers on diabetes research~\cite{sen2008collective} with 44K
citation edges and 3 disease-type labels.
Structurally analogous to ogbn-arxiv but in the biomedical domain,
testing whether citation-graph reasoning transfers across scientific
disciplines.
\textit{License.} PubMed graph (Sen et al., 2008): publicly
released for research; PubMed abstracts: \textbf{U.S.\ Government
public-domain} (NLM courtesy notice required when redistributed).

\paragraph{WikiCS (encyclopedia).}
10{,}893 Wikipedia articles on computer science topics connected by
276K hyperlink edges across 10 categories~\cite{mernyei2020wiki}.
Article text is rich and well-structured, making SBERT verification
straightforward.
\textit{License.} WikiCS graph + features (Mernyei \&\ Cangea,
2020): \textbf{MIT}; underlying Wikipedia article text:
\textbf{CC-BY-SA 3.0}.

\paragraph{USPTO (patent citation).}
266K patents in the H04L (transmission of digital information) CPC
class connected by 2.0M citation edges across 19 CPC group
codes~\cite{leskovec2005graphs}.
Patent titles follow a formal style distinct from academic writing,
testing generalisation across text registers.
\textit{License.} USPTO patent grants and CPC classifications:
\textbf{U.S.\ Government public-domain}; the citation graph is
released by USPTO as bulk data.

\paragraph{Physics StackExchange (Physics Q\&A).}
234K physics questions connected by 105K related-question links
across 51 topic tags.
Conversational question titles are diverse even within the same tag,
requiring relaxed coherence thresholds.
\textit{License.} Physics SE post titles, bodies, and comments:
\textbf{CC-BY-SA 4.0} (per Stack Exchange Network terms; attribution
to the original author and a link back to the question are required
when redistributing individual posts).

\section{Detailed Task Specifications}
\label{app:tasks}

For each task we list the setup, the no-shortcut construction, the
ground-truth source, and (for T4/T5) the structured output format.

\subsection{T1: Masked Node Prediction (node level, description)}

\textbf{Setup.}
A subgraph centred on a hub node whose title is replaced with
\masked.
The model must infer what the masked node is about from its
neighbours' titles and the link structure.

\textbf{No-shortcut construction.}
The hub's title is suppressed; any correct answer must come from
neighbourhood context.

\textbf{Ground truth.}
The real abstract of the masked node (retrieved from dataset
metadata, no annotation).
Hard metric: SBERT-Sim $\geq$ threshold against abstract.

\textbf{Domain adaptation.}
Across domains, the prompt vocabulary changes (``paper'',
``product'', ``article'', ``patent'', ``question'') and SBERT
thresholds are calibrated to entity text length (papers 0.30,
products 0.25, articles 0.25, patents 0.25, questions 0.20).

\subsection{T2: Relational Description (edge level, description)}

\textbf{Setup.}
Given an edge $(u, v)$ and $u$'s local subgraph, describe the
relationship, namely what role $v$ plays in $u$'s context.
``They share a topic'' is insufficient.

\textbf{No-shortcut construction.}
The role depends on both nodes' structural positions; neither node's
title alone encodes it.

\textbf{Ground truth.}
Subject label of $v$ (hard anchor); SBERT against the domain
description of $v$'s category (soft anchor).

\subsection{T3: Theme Summarisation (graph level, description)}

\textbf{Setup.}
A cluster of same-label nodes connected by edges within the cluster.
The model must describe the collective theme.

\textbf{No-shortcut construction.}
The theme is a collective property; no single node's title states
the area-level theme.

\textbf{Ground truth.}
Dominant label of the cluster (hard anchor); SBERT against a
reference theme description (soft anchor).

\subsection{T4: Outlier Detection (graph level, comparison)}

\textbf{Setup.}
A topically coherent cluster plus one injected outlier from a
different category.
The outlier shares edges with the cluster (structurally embedded)
but its title embedding has cosine similarity $<0.40$ to the cluster
centroid (topically distant).
Node IDs are shuffled.
The model must output one line per node:
\begin{quote}\small
\texttt{Node 0: cluster member} \\
\texttt{Node 1: outlier} \\
\texttt{...} \\[4pt]
\texttt{Reasoning: [2--3 sentence explanation]}
\end{quote}

\textbf{No-shortcut construction.}
The outlier is structurally embedded but topically distant;
neither topology nor a single node's content resolves the task.

\textbf{Ground truth.}
Injected outlier's node ID (deterministic, zero annotation cost).
Hard metrics: per-node precision, recall, F1 from structured lines.
Soft metric: SBERT of explanation against outlier's hidden abstract.

\subsection{T5: Community Detection (graph level, comparison)}

\textbf{Setup.}
A shuffled graph of nodes from 2--3 distinct categories.
Node IDs are interleaved across communities so that position encodes
no information.
The model must output:
\begin{quote}\small
\texttt{Node 0: Community 0} \\
\texttt{Node 1: Community 1} \\
\texttt{...} \\[4pt]
\texttt{Reasoning:} \\
\texttt{Community 0: [description]} \\
\texttt{Community 1: [description]}
\end{quote}

\textbf{No-shortcut construction.}
The partition is a collective property; no single node's text
reveals which other nodes belong to its group.

\textbf{Ground truth.}
Subject-label partition (deterministic, zero annotation cost).
Hard metrics: ARI, NMI, purity, community-count accuracy.
Soft metric: per-community SBERT against the category description.

\section{Sample Pool Generation}
\label{app:candidate_pool}

This section describes the pre-quality-control data pool.
\textbf{66{,}000 candidate samples} (6 domains $\times$ 5 tasks
$\times$ 2{,}200) are the raw input to the four-layer pipeline that
filters them down to the released 42{,}000
(Section~\ref{sec:quality}). At this stage no LLM judge, human
annotator, or deduplicator has touched the samples.

\paragraph{Domain coverage.}
The pool draws from the six text-attributed graphs listed in
Table~\ref{tab:domains}: \texttt{ogbn-arxiv}, \texttt{ogbn-products},
\texttt{PubMed}, \texttt{WikiCS}, \texttt{USPTO}, and
\texttt{Physics~SE}.

\paragraph{Subgraph sampling.}
For each domain we extract 2-hop ego-graphs centred on randomly
sampled hub nodes (in-degree $\geq 3$ with valid node text), with
up to 10 neighbours per hop. The resulting ego-graph has at most
$1+10+10\cdot 10 = 111$ nodes. The sampler enforces three
properties:
\begin{itemize}\itemsep1pt
\item \textbf{Per-cell quota.} For each (domain, task) cell we
  continue sampling until 2{,}200 valid candidates accumulate, so
  the pool contains exactly $6 \times 5 \times 2{,}200 = 66{,}000$
  samples regardless of underlying graph size.
\item \textbf{Cross-task consistency.} All five tasks for a given
  domain draw from the \emph{same} ego-graph pool of that domain;
  this lets us audit and cap on a per-cell basis (e.g.\ deduplicate
  ego-graphs before splitting) without reshuffling the
  cell--cell allocation.
\item \textbf{No leakage of dataset labels.} Models receive node
  \emph{titles} only. Abstracts (arxiv, PubMed), product
  descriptions, and dataset class labels are withheld for ground
  truth construction; they are not visible in any prompt at any
  stage.
\end{itemize}

\paragraph{Per-task ground-truth construction.}
Each ego-graph is then converted into one prompt per task using
deterministic rules:
\textbf{T1} masks the ego-centre's title and the gold answer is
its dataset category; \textbf{T2} sub-samples a connected and a
disconnected node pair from the ego-graph and asks for an edge-or-no-edge
verdict; \textbf{T3} uses the dominant ego-graph category as gold
theme; \textbf{T4} injects a hub from a different category as the
outlier; \textbf{T5} partitions the ego-graph by node category and
asks the model to recover the partition. These rules are pure
graph operations (no model is involved at this stage), so the gold
answers are fully reproducible from the raw graph.

\paragraph{Domain-adaptive prompting.}
Task prompts use domain-specific vocabulary via a
\texttt{DOMAIN\_TERMS} registry that maps each domain to its
natural language:
\begin{quote}\small
ogbn-arxiv: \textit{paper, papers, citation graph, cites} \\
ogbn-products: \textit{product, products, co-purchase graph,
co-purchased with} \\
PubMed: \textit{paper, papers, clinical citation graph, cites} \\
USPTO: \textit{patent, patents, citation graph, cites} \\
WikiCS: \textit{article, articles, hyperlink graph, links to} \\
Physics~SE: \textit{question, questions, related-question graph,
links to a related question}
\end{quote}
This rewrites surface vocabulary so the prompt reads naturally for
the domain (``this patent citation graph contains 7 patents'', not
``this graph contains 7 papers'') without changing task logic or
gold answers.

\paragraph{Per-task prompt templates.}
The five tasks share a single prompt skeleton: a graph header
(domain-specific noun, edge type, and node listing), a question
parameterised by the domain registry, and an answer-format block.
\textsc{\{noun\}}, \textsc{\{noun-plural\}}, \textsc{\{graph-kind\}},
\textsc{\{edge\}}, \textsc{\{topic\}}, and \textsc{\{theme\}} are
filled from \texttt{DOMAIN\_TERMS} above. For each task we show the
template once; the same template is used in all six domains.

\begin{figure*}[t]
\centering
\begin{promptblock}{Prompt template: T1 -- Masked Node Prediction}
\scriptsize\ttfamily
<graph> The \{graph-kind\} contains \{N\} \{noun-plural\}. Node \{hub\} is the central \{noun\} and its title is masked. Based on the graph structure and the titles of its \{deg\} neighbours, what is the masked \{noun\} most likely about? \\
Provide your answer in the format: \\
Answer: <one sentence stating the \{topic\}> \\
Reasoning: <2-4 sentences explaining why>
\end{promptblock}
\begin{promptblock}{Prompt template: T2 -- Relational Description}
\scriptsize\ttfamily
<graph> Consider Node \{a\} and Node \{b\} in this \{graph-kind\}. First, determine whether they have a \{edge\}. Then explain the relationship (or lack thereof). \\
Provide your answer in the format: \\
Answer: <one sentence stating whether the \{edge\} exists> \\
Reasoning: <2-3 sentences explaining why>
\end{promptblock}
\begin{promptblock}{Prompt template: T3 -- Theme Summarisation}
\scriptsize\ttfamily
<graph> This \{graph-kind\} contains \{N\} \{noun-plural\} connected by \{E\} \{edge\}s. What is this group of \{noun-plural\} collectively about? \\
Provide your answer in the format: \\
Answer: <one sentence naming the common \{theme\}> \\
Reasoning: <2-4 sentences explaining why>
\end{promptblock}
\begin{promptblock}{Prompt template: T4 -- Outlier Detection}
\scriptsize\ttfamily
<graph> This \{graph-kind\} contains \{N\} \{noun-plural\} connected by \{E\} \{edge\}s. Most \{noun-plural\} share a coherent \{theme\}; exactly ONE \{noun\} is the semantic outlier (its \{topic\} does not fit). Identify the outlier and explain why. \\
Provide your answer in the format: \\
Answer: \\
Node 0: <outlier | not outlier> \\
Node 1: <outlier | not outlier> \\
\dots\ (one line per node, exactly one must be 'outlier') \\
Reasoning: <2-3 sentences explaining why the outlier does not fit>
\end{promptblock}
\begin{promptblock}{Prompt template: T5 -- Community Detection}
\scriptsize\ttfamily
<graph> This \{graph-kind\} contains \{N\} \{noun-plural\} from \{K\} distinct communities. Partition the nodes into \{K\} communities labelled Community 0..\{K-1\}. Describe each community. \\
Provide your answer in the format: \\
Answer: \\
Node 0: Community <k> \\
Node 1: Community <k> \\
\dots\ (one line per node; every node assigned to exactly one community) \\
Reasoning: \\
Community 0: <2-3 sentence description of its \{theme\}> \\
Community 1: <2-3 sentence description of its \{theme\}> \\
\dots\ (one block per community)
\end{promptblock}
\caption{Per-task prompt templates. The five tasks share a single skeleton (graph header + question + answer-format block); only the question and answer-format change. Domain-specific tokens (\textsc{\{noun\}}, \textsc{\{edge\}}, etc.) are filled from \texttt{DOMAIN\_TERMS}.}
\label{fig:task_prompts}
\end{figure*}

The literal token \texttt{<graph>} is replaced with a plain-text
node listing (one line per node: \texttt{Node \{i\}: \{title\}})
followed by an edge listing (\texttt{Node \{a\} \{edge-verb\} Node
\{b\}}). Note that \emph{no} prompt ever contains the dataset class
label, the masked node's own title, the outlier's identity, or the
gold partition. Those live only in \texttt{ground\_truth} and are
never seen by the generator that drafts the placeholder reasoning,
nor by any test-time model.

\section{Data Quality Control Details}
\label{app:verification}

The four layers split into two jobs. \textbf{Layers 1 to 3 audit the
gold \emph{reasoning}} under a shared five-code failure vocabulary
(\textsc{C1~Fact}, \textsc{C2~Consistency},
\textsc{C3~Label-Integrity}, \textsc{C4~Logic}, \textsc{C5~Evidence})
distilled from the agentic audit (Section~\ref{sec:quality}), so
scripted, model, and human verdicts are directly comparable. Layer~1
catches mechanical defects with regex. Layer~2 adds two open-weight
70B judges for the semantic codes. Layer~3 calibrates Layer~2
against humans. \textbf{Layer~4 audits the \emph{observed graph}},
deduplicating any sample whose canonical identity already appears in
the release.

\subsection{Layer 1: Agent-audit-grounded rule-based filtering}
\label{app:release_gate_l1}

Layer~1 catches deterministic defects with no LLM and no GPU.
Each rule was added in response to a recurrent failure mode that
the agentic audit identified on a held-out diagnostic subset, then
operationalised as a regex or parser so that it can be applied
exhaustively to all 66{,}000 candidates.

\paragraph{Codes Layer~1 covers.}
Layer~1 covers the mechanically-detectable subset of the five-code
vocabulary:

\begin{itemize}
\item \textbf{C1 (\textsc{Fact}).}
  Every \texttt{Node $N$} reference in the gold reasoning must satisfy
  $0 \leq N < |V|$.
  We use a case-sensitive regex
  (\texttt{\textbackslash bNode\textbackslash s+\textbackslash d+\textbackslash b}),
  so that lowercase mentions inside literal placeholder titles such as
  \texttt{`Item (node 821252)'} (common in the
  \texttt{ogbn-products} subset where some products lack a textual
  title) are not flagged as hallucinations.

\item \textbf{C2 (\textsc{Consistency}, mechanical only).}
  For T2 we extract the polarity of the answer's edge claim (a
  positive verb cluster such as \texttt{have a citation/edge/inter\-action}
  versus a negative cluster such as \texttt{no edge / do not have / are not connected})
  and compare with the ground-truth flag \texttt{has\_edge}.
  For T4 we parse the unique \texttt{Node $N$: outlier} line and
  compare with the ground-truth outlier id.
  Semantic C2 violations (e.g.\ a reasoning that concludes
  ``no direct citation'' while the answer asserts an edge) require
  natural-language understanding and are deferred to Layer~2.

\item \textbf{C3 (\textsc{Label-Integrity}).}
  The answer string and the ground-truth label must be well-formed:
  no unbalanced quotes, no leading-quote fragments such as
  \texttt{`"Patio'} (a CSV-truncation artefact of
  \texttt{`"Patio, Lawn \& Garden"'}), no placeholder tokens
  (\texttt{---}, \texttt{???}, \texttt{[MASKED]}), and no degenerate
  labels shorter than three characters.

\item \textbf{Format compliance (T4, T5).}
  T4 must contain exactly one \texttt{Node $N$: outlier} line and
  exactly $|V|-1$ \texttt{Node $N$: not outlier} lines, with each
  local id $0,\ldots,|V|-1$ appearing on exactly one line.
  T5 must contain one \texttt{Node $N$: Community $X$} line per node
  and the number of distinct communities must agree with the
  ground-truth partition.
\end{itemize}

\textbf{C4 (\textsc{Logic})} and \textbf{C5 (\textsc{Evidence})} are
inherently semantic and live in Layer~2.

\paragraph{Cost and throughput.}
Layer~1 is $\sim$250 lines of Python regex that runs over the full
66{,}000-sample pool in under five minutes on a CPU at zero cost.

\paragraph{Results.}
Of 66{,}000 candidates, Layer~1 admits 65{,}328 (98.98\%) and rejects
672 (Table~\ref{tab:l1_results}).
The dominant rejection cause is C3 (671 samples, 99.9\% of all
rejections, concentrated on T1 and T3) and traces to a single root
cause: a CSV-parsing bug in the upstream \texttt{ogbn-products} label
loader that truncates labels containing literal commas
(\texttt{`"Patio, Lawn \& Garden"'} $$ to $$ \texttt{`"Patio'}),
producing an unbalanced leading quote in both the gold label and any
answer that quotes it.
After the regex fix that suppresses lowercase node mentions inside
literal placeholder titles such as \texttt{`Item (node 821252)'},
only two C1 cases survive across the full 66K release: genuine
out-of-range reasoning references that do warrant exclusion.
T5 has zero Layer-1 rejections after the fix, reflecting the fact
that T5 templates are generated by a structured serialiser rather
than free-text reasoning.

\begin{table}[h]
\centering
\caption{Layer-1 data-quality-control results on the 66{,}000-sample
  candidate release (5 tasks $\times$ 6 domains $\times$ 2{,}200).
  C3 dominates rejections through a single upstream CSV-truncation
  defect on \texttt{ogbn-products}; the residual C1/C2/format
  failures are vanishingly rare ($\leq 2$ per task).}
\label{tab:l1_results}
\small
\begin{tabular}{@{}lrrr@{}}
\toprule
\textbf{Task} & \textbf{N} & \textbf{Rejected} & \textbf{Qualified} \\
\midrule
T1 (Masked Node)               & 13{,}200 & 196 & 98.52\% \\
T2 (Relational Semantics)      & 13{,}200 &  95 & 99.28\% \\
T3 (Theme Summarization)       & 13{,}200 & 226 & 98.29\% \\
T4 (Outlier Detection)         & 13{,}200 & 155 & 98.83\% \\
T5 (Community Detection)        & 13{,}200 &   0 & 100.00\% \\
\midrule
\textbf{Total}                 & 66{,}000 & 672 & \textbf{98.98\%} \\
\bottomrule
\end{tabular}
\end{table}

\subsection{Layer 2: Dual open-source 70B judges}
\label{app:release_gate_l2}

Layer~2 catches the semantic violations Layer~1 cannot see:
ungrounded entity references that survive the C1 regex
(e.g.\ a hallucinated \emph{title} matching no node), reasoning--answer
self-contradictions (C2-semantic), illogical inference steps (C4),
and reasoning that cites the wrong neighbours for the queried claim
(C5).

\paragraph{Judge models.}
We deploy two open-weight 70B-class judges,
\textbf{Llama-3.1-70B-Instruct} and \textbf{Qwen-2.5-72B-Instruct},
both AWQ-INT4~\cite{lin2024awq} and served via vLLM with the
AWQ-Marlin kernel, greedy decoding, and a 300-token generation
budget. Each AWQ checkpoint fits in $\sim$40\,GB, so a single RTX
PRO 6000 Blackwell GPU (97\,GB) hosts one judge instance and a fleet
dispatches samples in parallel. Open weights make the verdicts
reproducible by any third party with comparable hardware.

\paragraph{Prompt and aggregation.}
Both judges receive the identical system prompt enumerating
\textsc{C1}--\textsc{C5}, plus a user message that supplies the gold
answer, the gold reasoning, the ground-truth structure, and the
local graph context (node titles truncated to 8\,000 characters).
The judge emits a single-line JSON verdict
\texttt{\{qualified, fail\_flags, rationale\}}.
We aggregate via \texttt{all\_pass}: a sample is admitted only if
both judges return \texttt{qualified=true}.
Figure~\ref{fig:l2_prompt} reproduces the verbatim prompt template
used for both judges (no domain- or task-specific
parameter-tuning beyond the \texttt{\{task\_desc\}} string that
expands to \texttt{Masked Node Prediction}, \texttt{Relational
Semantics}, etc., as listed in Appendix~\ref{app:tasks}).

\begin{figure*}[t]
\centering
\begin{promptblock}{System prompt (identical for both judges)}
\scriptsize
You are a strict release-quality auditor for the GraphInfer-Bench
benchmark. You will judge exactly ONE sample. Decide whether it is
QUALIFIED for the public release. Be conservative: when in doubt,
FAIL.

A sample is QUALIFIED only if \textbf{all five} checks pass:

\textbf{C1 FACT}: every node ID / title / protein / paper cited in
the reasoning appears in the graph context above. No fabricated
neighbours.

\textbf{C2 CONSISTENCY}: the conclusion of the reasoning agrees
with the answer. Specifically, reject if the reasoning states ``no
edge exists'', ``no direct citation'', or ``no interaction'' while
the answer asserts the edge is present (or vice versa). Reject if
the reasoning concludes one subject label and the answer states a
different one.

\textbf{C3 LABEL\_INTEGRITY}: the answer string is well-formed:
no CSV-truncation artefacts (\eg\ \texttt{\textquotedbl Patio} is a
fragment of \texttt{\textquotedbl Patio, Lawn \& Garden\textquotedbl}
and is invalid), no stray quotes, no obviously wrong category names.

\textbf{C4 LOGIC}: each step of the reasoning follows from the
previous step or from the graph context. Reject unwarranted leaps
and post-hoc justifications that don't engage with the visible
neighbours.

\textbf{C5 EVIDENCE}: the reasoning uses the RIGHT neighbours /
edges to support its claim. For T2, the reasoning must justify the
queried pair (not a different edge). For T4, the cited cluster
signal must actually differentiate the outlier. For T5, each
community claim must cite nodes assigned to that community.

Output \textbf{exactly one} JSON object on a single line, nothing
else:
\texttt{\{"qualified": true|false, "fail\_flags":
[...subset of "C1","C2","C3","C4","C5"...], "rationale": "..."\}}
\end{promptblock}

\begin{promptblock}{User message template (per sample)}
\scriptsize\ttfamily
Domain: \{domain\}\\
Task:\ \ \ \{task\} (\{task\_desc\})\\[2pt]
Graph context (node titles):\\
---\\
\{node\_titles\}\\
---\\[2pt]
Ground truth (hidden from the model):\\
---\\
\{gt\}\\
---\\[2pt]
Gold answer (will be released as the reference label):\\
---\\
\{answer\}\\
---\\[2pt]
Gold reasoning (will be released as the reference rationale):\\
---\\
\{reasoning\}\\
---
\end{promptblock}
\caption{Layer-2 judge prompt. Both Llama-3.1-70B-AWQ and
Qwen-2.5-72B-AWQ receive an identical \emph{system prompt}
(top, defining \textsc{C1}--\textsc{C5}) and a per-sample
\emph{user message} (bottom). Decoding is greedy
(temperature~$=0$); generation budget 300 tokens.}
\label{fig:l2_prompt}
\end{figure*}

\paragraph{Throughput.}
We dispatch the 65{,}328 Layer-1-passing samples across a
16-GPU fleet (4 servers $\times$ 1, 3, 5, 6 RTX PRO 6000
Blackwells), with deterministic per-judge sharding:
$\mathrm{md5}(\text{sample\_id})\bmod{N_\mathrm{judge}} = k$
assigns each sample to one of $N_\mathrm{judge}=8$ shards per judge.
Aggregate throughput peaks at $\sim$\,30 verdicts$/$s
($\sim$15$/$s per judge) once vLLM prefix caching warms on the
shared system prompt; the full 130{,}656-call run completes in
$\sim$1.7 wall hours.

\paragraph{Results.}
Of the 65{,}328 Layer-1-passing samples, both judges return verdicts
on 65{,}164; the missing $164$ trace to transient parser errors and
are dropped.
Llama qualifies 97.9\% of the samples it sees;
Qwen qualifies 95.7\%. Qwen is the systematically stricter judge.
The \texttt{all\_pass} aggregation (Table~\ref{tab:l2_results})
admits \textbf{62{,}010} samples (95.2\% of both-judged samples,
94.9\% of the L1-passing set, 93.9\% of the master 66{,}000)
and drops the rest: 1{,}054 (1.6\%) on which both judges agree to
reject and 2{,}100 (3.2\%) on which the two judges disagree.
We treat any single-judge rejection as a rejection of the whole
sample.

\paragraph{Inter-judge agreement.}
On the 65{,}164 dual-judged sample pairs (6 domains, post
string-db drop), raw agreement is 96.8\% and Cohen's
$\kappa = 0.486$, substantial agreement, well above chance.
Disagreements are highly asymmetric: 84\% of them are
Llama-pass-Qwen-reject (the stricter judge catching what the lenient
one missed), versus only 16\% the other way. This is the expected
signature of two independent judges on a clean dataset where most
failures are subtle.

\paragraph{Failure-flag distribution (6 domains).}
Llama flags $1{,}401$ samples and Qwen flags $2{,}807$. A single
rejection may carry multiple flags. Both judges share the
\textsc{C2}~$\succ$~\textsc{C1}~$\succ$~\textsc{C4}~$\succ$~\textsc{C5}
ordering. Qwen: \textsc{C2}~$2{,}452$ (87.4\%), \textsc{C3}~$879$
(31.3\%), \textsc{C1}~$815$ (29.0\%), \textsc{C4}~$616$ (21.9\%),
\textsc{C5}~$567$ (20.2\%). Llama: \textsc{C2}~$742$ (53.0\%),
\textsc{C1}~$444$ (31.7\%), \textsc{C4}~$249$ (17.8\%),
\textsc{C5}~$112$ (8.0\%). Qwen is calibrated tighter, especially on
\textsc{C2} self-contradictions.

\begin{table}[h]
\centering
\caption{Layer-2 data-quality-control results on the 65{,}328 Layer-1-passing
  samples (5 tasks $\times$ 6 domains, computed after the string-db
  drop). \texttt{all\_pass} admits a sample only if both
  Llama-3.1-70B-AWQ and Qwen-2.5-72B-AWQ return
  \texttt{qualified=true}; both-reject and inter-judge disagreement
  outcomes are dropped from the release.}
\label{tab:l2_results}
\small
\begin{tabular}{@{}lrr@{}}
\toprule
\textbf{Outcome} & \textbf{Count} & \textbf{\%} \\
\midrule
Both qualify (\texttt{all\_pass}, ship) & 62{,}010 & 95.2\% \\
Both reject (drop)                      &  1{,}054 &  1.6\% \\
\quad agreement subtotal                & 63{,}064 & 96.8\% \\
Llama-pass / Qwen-reject                &  1{,}753 &  2.7\% \\
Qwen-pass / Llama-reject                &     347 &  0.5\% \\
\quad disagreement subtotal             &  2{,}100 &  3.2\% \\
\midrule
Cohen's $\kappa$ (Llama-70B vs.\ Qwen-72B)      & \multicolumn{2}{r}{$0.486$} \\
\bottomrule
\end{tabular}
\end{table}

\subsection{Layer 3: Human \texorpdfstring{$\kappa$}{kappa} calibration}
\label{app:release_gate_l3}

Layer~3 certifies that Layer-2 verdicts are trustworthy. It does
not adjudicate individual samples. Two annotators independently
rendered \textsc{C1} to \textsc{C5} verdicts on a 300-sample
stratified subset (10 per task-domain cell, sampled to match the L2
fail-flag distribution). We require inter-annotator Cohen's $\kappa
\geq 0.6$ on binary ship/drop and L2-vs-human-consensus agreement
$\geq 95\%$ on the qualified/reject label. \textbf{Both thresholds
are met} ($\kappa = 0.606$, agreement $= 95.2\%$) with zero L2
silent-pass errors. The L2 \texttt{all\_pass} cohort therefore ships
as-is.

\subsection{Layer 4: Construction-time leakage prevention}
\label{app:release_gate_l4}

The first three layers police whether each individual sample is
\emph{correct}. Layer~4 polices whether the release as a \emph{set}
is free of leakage: no single ego-centre, edge, cluster, outlier
pair, or sub-graph may appear in more than one of the public
train/val/test splits, and no sample is duplicated within a split.
Because we deduplicate \emph{before} the random split is drawn,
the leakage-free property is a construction-time guarantee, not a
post-hoc cleanup.

\paragraph{Per-task canonical identity.}
Identities are computed from \emph{global} graph node ids (the ids in
the upstream graph, not the per-ego-graph local ids), so two samples
that re-numbered the same nodes still collide:

\begin{itemize}
\item \textbf{T1 (Masked Node)}: the masked ego-centre's
  global node id.
\item \textbf{T2 (Relational Semantics)}: the unordered pair of
  global node ids of the two endpoints
  $\{\textsf{global}(\textsf{node\_a\_id}), \textsf{global}(\textsf{node\_b\_id})\}$.
\item \textbf{T3 (Theme Summarization)}: the frozenset of all
  global node ids in the ego-graph (the cluster signature).
  Additionally, any pair of T3 samples whose neighbour-set Jaccard
  similarity is $\geq 0.5$ is treated as a near-duplicate; the
  later-drawn sample of each such pair is dropped.
\item \textbf{T4 (Outlier Detection)}: the (outlier global id,
  frozenset of cluster global ids) tuple, so any
  outlier-vs-same-cluster repeat is caught.
\item \textbf{T5 (Community Detection)}: the frozenset of all
  global node ids in the ego-graph (the sub-graph signature).
\end{itemize}

For each (domain, task) cell we keep one sample per identity using
a deterministic tie-break (lexicographically smallest sample id).

\paragraph{Effect on the release.}
Layer~4 drops $1{,}132$ samples by exact identity collision and an
additional $1{,}360$ T3 near-duplicates (Jaccard $\geq 0.5$),
$2{,}492$ in total, $4.0\%$ of the L2-passing set
(Table~\ref{tab:l4_results}). The final public release contains
\textbf{59{,}681 samples} on which the train/val/test splits are
then drawn uniformly at random, with split membership disjoint by
construction.
Layer~4 hits hardest on \texttt{physics-se} (T3 alone loses 780
samples) and \texttt{ogbn-arxiv} T3 (449 samples), reflecting that
small sub-graph diameters in those domains make near-isomorphic
ego-graphs more likely under random sampling.

\begin{table}[h]
\centering
\caption{Layer-4 deduplication results, per task. ``Exact dup''
  counts canonical-identity collisions; ``T3 near-dup'' counts T3
  sample pairs whose neighbour-set Jaccard is $\geq 0.5$ (only the
  later-drawn member of each pair is dropped). The kept column is
  the post-L4 contribution per task; their sum is the public
  release set.}
\label{tab:l4_results}
\small
\begin{tabular}{@{}lrrrr@{}}
\toprule
\textbf{Task} & \textbf{In} & \textbf{Exact dup} & \textbf{T3 near-dup} & \textbf{Kept} \\
\midrule
T1 (Masked Node)               & 11{,}911 &      0 &     -- & 11{,}911 \\
T2 (Relational Semantics)      & 11{,}735 &     97 &     -- & 11{,}638 \\
T3 (Theme Summarization)       & 12{,}790 &    592 & 1{,}360 & 10{,}838 \\
T4 (Outlier Detection)         & 12{,}947 &    396 &     -- & 12{,}551 \\
T5 (Community Detection)        & 12{,}790 &     47 &     -- & 12{,}743 \\
\midrule
\textbf{Total}                 & 62{,}173 & 1{,}132 & 1{,}360 & \textbf{59{,}681} \\
\bottomrule
\end{tabular}
\end{table}

\begin{figure}[h]
\centering
\includegraphics[width=0.96\linewidth]{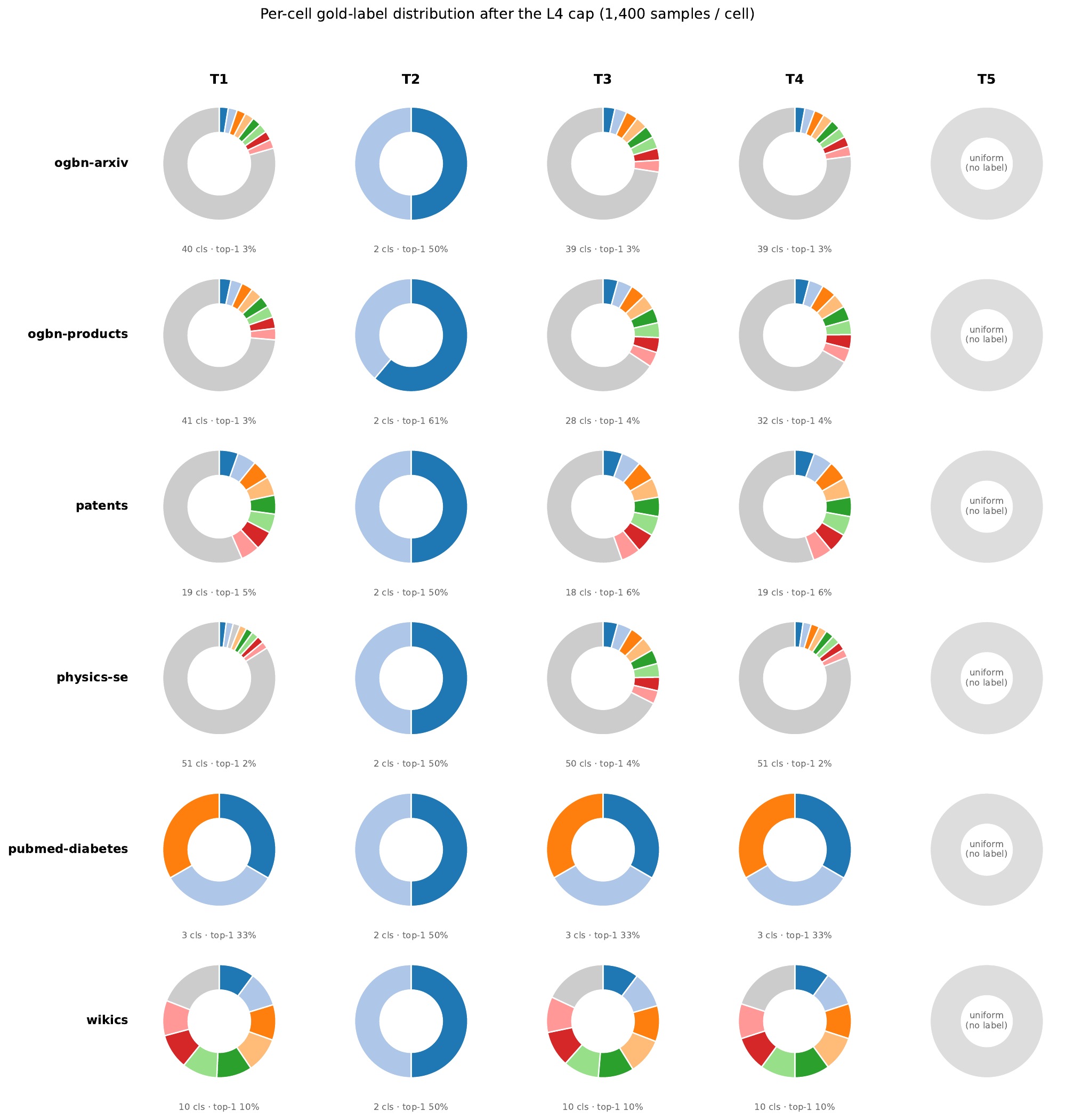}
\caption{Per-cell gold-label distribution after the L4 cap
  (1{,}400 samples / cell). Each donut is one (domain, task)
  cell; wedges are the top-8 gold classes drawn in fixed order,
  and the grey \emph{other} ring collects the long tail.
  Captions below each donut report (\#classes,
  top-1 share). The visual signature confirms that the stratified
  sampler kept every cell close to its uniform floor $1/K$ rather
  than concentrating mass in one class.}
\label{fig:label_pies}
\end{figure}

\paragraph{Cap and per-cell label balance.}
After dedup, every (domain, task) cell is capped to 1{,}400
samples and split 1{,}000 train / 100 val / 300 test. To prevent
the cap from amplifying any pre-existing label skew, we sample
\emph{stratified on the gold balance label} (\textsc{T1}: subject
label of the masked node; \textsc{T2}: \texttt{has\_edge} flag;
\textsc{T3}: cluster dominant label; \textsc{T4}: cluster subject
label; \textsc{T5}: uniform random, since it has no single gold label
per sample), and force exact split sizes by moving overflow into
the next under-quota split. The resulting public release contains
\textbf{42{,}000} samples (30{,}000 train / 3{,}000 val / 9{,}000
test). Figure~\ref{fig:label_pies} shows the per-cell label
distribution as a $6 \times 5$ grid of donut charts: each donut
breaks the 1{,}400 samples of one (domain, task) cell into the
top-8 gold classes plus a grey \emph{other} slice for the long
tail. Below each donut we annotate the number of distinct
labels and the top-1 share. T2 lands at $\sim$50/50
(has-edge / no-edge) by construction; pubmed-diabetes lands at
$\sim$33\% top-1 (3 disease classes); wikics at $\sim$10\%
(10 categories); the larger fine-grained taxonomies of
ogbn-arxiv, ogbn-products, patents and physics-se sit at
2--6\% top-1 with a long tail of small slices. T5 has no per-sample
gold label, so its column is shown as a single uniform slice.

\section{Per-Domain GNN Reference}
\label{app:per_domain_gnn}

Table~\ref{tab:domain_gnn} reports the per-cell mean$\pm$std of three
random seeds (42, 43, 44) for the GNN structural reference
(GCN, GAT, GraphSAGE) trained on \texttt{data/splits\_v4} (1{,}000
train / 100 val / 300 test per cell). The aggregate domain-averaged
numbers are the GNN block of Table~\ref{tab:main}. T1, T2, and T3
report hard-label accuracy, T4 reports Hit@1 of the
outlier-vs-anchor margin, and T5 reports the Normalized Mutual
Information of the predicted clustering against the gold community
labels.

\begin{table*}[h]
\centering
\caption{Per-domain results of the GNN structural reference (3 seed).
  All metrics are higher-is-better. Each cell is mean$\pm$std on the
  300-sample test split.}
\label{tab:domain_gnn}
\small
\renewcommand{\arraystretch}{1.1}
\setlength{\tabcolsep}{4pt}
\begin{tabular}{@{}ll ccccc@{}}
\toprule
\textbf{Model} & \textbf{Domain}
  & \textbf{T1 Acc.} & \textbf{T2 Acc.}
  & \textbf{T3 Acc.} & \textbf{T4 H@1} & \textbf{T5 NMI} \\
\midrule
  \multirow{6}{*}{GCN}
  & ogbn-arxiv      & 0.538$\pm$0.006 & 0.743$\pm$0.005 & 0.831$\pm$0.008 & 0.110$\pm$0.000 & 0.149$\pm$0.007 \\
  & ogbn-products   & 0.623$\pm$0.007 & 0.833$\pm$0.011 & 0.874$\pm$0.004 & 0.087$\pm$0.007 & 0.209$\pm$0.018 \\
  & patents         & 0.467$\pm$0.003 & 0.852$\pm$0.013 & 0.870$\pm$0.012 & 0.076$\pm$0.004 & 0.240$\pm$0.003 \\
  & physics-se      & 0.371$\pm$0.020 & 0.899$\pm$0.007 & 0.806$\pm$0.006 & 0.090$\pm$0.010 & 0.164$\pm$0.006 \\
  & pubmed-diabetes & 0.856$\pm$0.002 & 0.944$\pm$0.004 & 0.981$\pm$0.003 & 0.131$\pm$0.018 & 0.324$\pm$0.012 \\
  & wikics          & 0.792$\pm$0.018 & 0.826$\pm$0.016 & 0.992$\pm$0.002 & 0.208$\pm$0.011 & 0.346$\pm$0.002 \\
\midrule
\multirow{6}{*}{GAT}
  & ogbn-arxiv      & 0.552$\pm$0.009 & 0.510$\pm$0.009 & 0.836$\pm$0.008 & 0.293$\pm$0.005 & 0.232$\pm$0.002 \\
  & ogbn-products   & 0.623$\pm$0.007 & 0.620$\pm$0.003 & 0.867$\pm$0.012 & 0.080$\pm$0.012 & 0.280$\pm$0.003 \\
  & patents         & 0.476$\pm$0.007 & 0.582$\pm$0.012 & 0.862$\pm$0.007 & 0.124$\pm$0.012 & 0.249$\pm$0.001 \\
  & physics-se      & 0.378$\pm$0.011 & 0.522$\pm$0.012 & 0.814$\pm$0.011 & 0.222$\pm$0.004 & 0.176$\pm$0.001 \\
  & pubmed-diabetes & 0.847$\pm$0.012 & 0.729$\pm$0.008 & 0.969$\pm$0.002 & 0.434$\pm$0.011 & 0.315$\pm$0.015 \\
  & wikics          & 0.804$\pm$0.018 & 0.566$\pm$0.007 & 0.989$\pm$0.004 & 0.213$\pm$0.000 & 0.353$\pm$0.000 \\
\midrule
\multirow{6}{*}{GraphSAGE}
  & ogbn-arxiv      & 0.529$\pm$0.008 & 0.605$\pm$0.016 & 0.812$\pm$0.006 & 0.393$\pm$0.007 & 0.255$\pm$0.003 \\
  & ogbn-products   & 0.619$\pm$0.017 & 0.689$\pm$0.003 & 0.876$\pm$0.002 & 0.179$\pm$0.006 & 0.326$\pm$0.000 \\
  & patents         & 0.450$\pm$0.024 & 0.699$\pm$0.004 & 0.853$\pm$0.010 & 0.201$\pm$0.019 & 0.300$\pm$0.000 \\
  & physics-se      & 0.353$\pm$0.008 & 0.792$\pm$0.018 & 0.819$\pm$0.004 & 0.453$\pm$0.011 & 0.273$\pm$0.004 \\
  & pubmed-diabetes & 0.866$\pm$0.004 & 0.833$\pm$0.005 & 0.959$\pm$0.004 & 0.760$\pm$0.019 & 0.792$\pm$0.028 \\
  & wikics          & 0.811$\pm$0.010 & 0.720$\pm$0.013 & 0.989$\pm$0.002 & 0.486$\pm$0.011 & 0.797$\pm$0.001 \\
\bottomrule
\end{tabular}
\renewcommand{\arraystretch}{1.0}
\end{table*}

\section{Failure Case Analysis}
\label{app:failure_cases}

To complement the aggregate metrics in Table~\ref{tab:main} and the per-domain
breakdown in Tables~\ref{tab:appendix_expanded_t123}--\ref{tab:appendix_expanded_t45},
we manually inspected
$\sim$1{,}200 predictions per (baseline, task) cell and characterise
the recurring failure modes below. None of the patterns are caused
by the rescorer. Per-baseline parser hit-rate is 89 to 100\% across
all five tasks, so the remaining gap between baseline and oracle
reflects \emph{model} weaknesses, not parsing.

\subsection{T5 (Community Detection): Mode Collapse}
\label{app:t5_collapse}

All seven baselines collapse to a near-degenerate partition: between
85\% and 99\% of nodes are assigned to a single community. The
remaining 1-15\% are typically scattered as singleton ``Community 1''
labels. The extreme case is GOFA on \texttt{ogbn-arxiv}, where 73/73
nodes are placed in Community 0. NMI is consequently dominated by
the chance-level baseline (NMI $\approx 0.02$-$0.07$ across baselines),
even though the parser correctly recovers the predicted partition in
$>$99\% of samples.

\subsection{T4 (Outlier ID): Plausible-but-Wrong Convergence}
\label{app:t4_convergence}

On the \texttt{ogbn-arxiv} sample shown in our spot-check, six of seven
baselines (LLaGA, GOFA, GraphGPT, GraphToken, RGLM, TEA-GLM) all
predict ``Node 28: outlier'' while the gold answer is Node 15. Each
baseline produces a distinct, well-formed reasoning paragraph
justifying its (wrong) choice -- an indication that they pick a
\emph{plausible-but-wrong} outlier rather than failing to engage with
the task. InstructGLM is the exception, mode-collapsing to ``Node 1''
in 664/1{,}050 (63\%) of T4 predictions regardless of input.

\subsection{LLaGA T4 -- Unfilled Template (11\%)}
\label{app:llaga_template}

A specific failure mode is observed only on LLaGA T4: in 11\% of
predictions the model emits the \emph{template placeholder verbatim},
e.g.\ ``Node 0: <outlier $\mid$ not outlier> Node 1: <outlier $\mid$ not
outlier> $\dots$''. The angle-bracket pattern is a literal copy from
the prompt, with no actual classification. These predictions are
treated as non-answers by our rescorer (counted in $n$ but excluded
from the H@1 numerator).

\subsection{T1/T3: Semantically-Related Wrong Labels}
\label{app:t1t3_wrong}

For T1 (subject classification) and T3 (cluster theme), the dominant
failure mode is selecting a \emph{semantically related but wrong}
label, e.g.\ predicting ``Computer Vision'' for a paper labeled
``Graphics'', or ``Machine Learning'' for ``Neural and Evolutionary
Computing''. The labels are drawn from the same parent ontology, so
SBERT-F1 remains high ($>0.9$ for e5-large in many cells) even when
hard accuracy is low. We deliberately do \emph{not} apply
label-hierarchy fuzzy matching: doing so would inflate scores
$\sim 5$-$10$pp without a principled stopping criterion. The
SBERT-F1 columns already capture this graceful-degradation signal.

\subsection{InstructGLM T5 -- Node Set Hallucination}
\label{app:instructglm_hallucination}

For T5 community detection, InstructGLM consistently emits assignments for
$\sim$500 nodes when the input ego-graph contains 50-100. The extra
nodes are syntactically valid (``Node 511: Community 0'') but
correspond to no actual node in the graph. Our rescorer takes the
intersection of predicted and reference node sets before computing
NMI, so the hallucinated nodes are ignored. The underlying behavior
suggests InstructGLM was trained on graphs of a different size
distribution than our 50-100 node ego-graph test set.

\subsection{Inference-Cost Issue: Infinite Generation}
\label{app:infgen}

We observe that LLaGA, GOFA, GraphGPT, GraphToken, RGLM, and TEA-GLM
do not terminate cleanly after the Answer/Reasoning block. In 80-99\%
of predictions, the model emits an end-of-sequence token (e.g.\
\texttt{</s>}), then resumes generation with a fresh
``ASSISTANT: I agree with the classification\dots'' prefix and writes
several more thousand characters. Median prediction length is
1-15\,KB across baselines (with InstructGLM as the only exception at
$<$1\,KB median). This does not affect rescoring -- our parser
extracts only the first valid Answer block -- but does waste
inference compute. A conservative re-evaluation with proper
\texttt{stop\_token\_ids} would cut inference cost roughly $5$-$10\times$
without changing reported numbers.

\subsection{Patents -- Frozen-LLM Output-Format Lock-in vs.\ True 19-way Difficulty}
\label{app:patents_finding}

The \texttt{patents} domain isolates a sharp distinction between
fine-tuned and frozen-LLM baselines that is invisible in any other
domain. The label space is 19 IPC subclasses under the H04L group, all
formatted as ``H04LN: <description>'' (e.g., ``H04L1: Error
detection/correction in transmission''). Training answers expose this
exact format ($\sim$54 examples per subclass).

\textbf{Graph-token baselines.} LLaGA, GraphToken, GraphGPT, RGLM,
GOFA, and TEA-GLM hover near 19-way chance (T1 hard accuracy
$\approx$ 4--6\%, vs.\ chance = 5.3\%). Inspecting their predictions
shows they emit free-form natural-language descriptors (``Cryptographic
protocols'', ``Data transmission systems'') rather than the IPC code
format. Because these methods freeze the LLM and only fine-tune a small
projector, the LLM never adapts its output distribution to the
domain-specific label vocabulary; the generated text reflects
pre-training priors instead of the trained label space.

\textbf{SFT baselines.} SFT with Llama-3-8B, Qwen-2-7B, and
Vicuna-7B all reproduce the IPC format on $>$99.5\% of patents
predictions (e.g., 288/288 for Llama-3, 282/283 for Qwen-2, 242/242
for Vicuna). Their hard accuracy of 35--45\% reflects the genuine
difficulty of distinguishing 19 closely-related H04L subclasses
(network security vs.\ traffic control vs.\ addressing vs.\ multimedia)
from a small set of cited-patent titles.

\textbf{Implication.} GraphInfer-Bench's \texttt{patents} cell separates
\emph{output-format adaptation failure} (frozen LLM can't emit
domain-specific label codes) from \emph{classification difficulty} (19
fine-grained subclasses with overlapping semantics). Both contribute to
the gap between graph-token baselines and SFT here; on natural-language
domains (ogbn-arxiv, pubmed-diabetes) only the latter applies.


\section{Per-Domain Results: Graph-Token Baselines}
\label{app:percell_graphtoken}
Per-(domain, task) hard accuracy and 3-SBERT triples (MiniLM-L6 / mpnet-base /
e5-large) for the seven graph-token alignment baselines. The aggregate row
of Table~\ref{tab:main} is the domain-averaged form of these cells.

\begin{table*}[h]
\centering
\caption{Per-domain results for every Graph-Token baseline (T1--T3). SBERT score reported across MiniLM-L6-v2 (M), mpnet-base-v2 (p), e5-large-v2 (E5) jointly. Each cell is mean$\pm$std across 3 seed.}
\label{tab:appendix_expanded_t123}
\scriptsize
\renewcommand{\arraystretch}{1.05}
\setlength{\tabcolsep}{2.5pt}
\begin{tabular}{@{}ll cc cc cc @{}}
\toprule
& & \multicolumn{2}{c}{\textbf{T1}} & \multicolumn{2}{c}{\textbf{T2}} & \multicolumn{2}{c}{\textbf{T3}} \\
\cmidrule(lr){3-4} \cmidrule(lr){5-6} \cmidrule(lr){7-8}
\textbf{Model} & \textbf{Domain} & Acc & SBERT (M/p/E5)  & Acc & SBERT (M/p/E5)  & Acc & SBERT (M/p/E5) \\
\midrule
  \multirow{6}{*}{LLaGA} & ogbn-arxiv         & 0.475\,$\pm$\,0.002 & 0.72/0.65/0.91 & 0.731\,$\pm$\,0.006 & 0.62/0.57/0.91 & 0.742\,$\pm$\,0.002 & 0.69/0.69/0.91 \\
   & ogbn-products      & 0.268\,$\pm$\,0.304 & 0.51/0.58/0.86 & 0.548\,$\pm$\,0.221 & 0.50/0.55/0.87 & 0.243\,$\pm$\,0.264 & 0.48/0.61/0.87 \\
   & patents            & 0.063\,$\pm$\,0.010 & 0.65/0.60/0.90 & 0.767\,$\pm$\,0.021 & 0.56/0.52/0.89 & 0.046\,$\pm$\,0.012 & 0.41/0.45/0.87 \\
   & physics-se         & 0.011\,$\pm$\,0.008 & 0.46/0.44/0.87 & 0.545\,$\pm$\,0.068 & 0.49/0.49/0.88 & 0.036\,$\pm$\,0.033 & 0.36/0.40/0.86 \\
   & pubmed-diabetes    & 0.380\,$\pm$\,0.180 & 0.67/0.67/0.91 & 0.717\,$\pm$\,0.137 & 0.60/0.61/0.90 & 0.482\,$\pm$\,0.133 & 0.59/0.60/0.89 \\
   & wikics             & 0.111\,$\pm$\,0.018 & 0.42/0.41/0.85 & 0.614\,$\pm$\,0.056 & 0.52/0.58/0.86 & 0.106\,$\pm$\,0.005 & 0.50/0.54/0.86 \\
\midrule
  \multirow{6}{*}{GraphToken} & ogbn-arxiv         & 0.161\,$\pm$\,0.192 & 0.53/0.45/0.89 & 0.682\,$\pm$\,0.030 & 0.54/0.56/0.90 & 0.365\,$\pm$\,0.441 & 0.48/0.46/0.88 \\
   & ogbn-products      & 0.037\,$\pm$\,0.009 & 0.42/0.48/0.84 & 0.685\,$\pm$\,0.066 & 0.47/0.55/0.86 & 0.091\,$\pm$\,0.010 & 0.36/0.46/0.85 \\
   & patents            & 0.045\,$\pm$\,0.026 & 0.60/0.56/0.89 & 0.752\,$\pm$\,0.003 & 0.57/0.52/0.89 & 0.062\,$\pm$\,0.012 & 0.45/0.48/0.88 \\
   & physics-se         & 0.021\,$\pm$\,0.004 & 0.52/0.46/0.89 & 0.863\,$\pm$\,0.020 & 0.46/0.44/0.89 & 0.043\,$\pm$\,0.023 & 0.35/0.42/0.87 \\
   & pubmed-diabetes    & 0.561\,$\pm$\,0.050 & 0.76/0.74/0.93 & 0.873\,$\pm$\,0.038 & 0.64/0.61/0.91 & 0.520\,$\pm$\,0.159 & 0.62/0.61/0.90 \\
   & wikics             & 0.161\,$\pm$\,0.042 & 0.55/0.51/0.88 & 0.741\,$\pm$\,0.017 & 0.54/0.54/0.87 & 0.151\,$\pm$\,0.046 & 0.51/0.53/0.88 \\
\midrule
  \multirow{6}{*}{GraphGPT} & ogbn-arxiv         & 0.025\,$\pm$\,0.016 & 0.40/0.29/0.87 & 0.604\,$\pm$\,0.112 & 0.50/0.61/0.89 & 0.045\,$\pm$\,0.007 & 0.29/0.26/0.86 \\
   & ogbn-products      & 0.032\,$\pm$\,0.002 & 0.42/0.49/0.84 & 0.572\,$\pm$\,0.237 & 0.44/0.52/0.86 & 0.060\,$\pm$\,0.014 & 0.32/0.47/0.84 \\
   & patents            & 0.041\,$\pm$\,0.018 & 0.65/0.60/0.90 & 0.730\,$\pm$\,0.017 & 0.58/0.53/0.89 & 0.013\,$\pm$\,0.001 & 0.43/0.46/0.87 \\
   & physics-se         & 0.015\,$\pm$\,0.007 & 0.53/0.48/0.88 & 0.785\,$\pm$\,0.066 & 0.47/0.47/0.89 & 0.052\,$\pm$\,0.007 & 0.37/0.41/0.86 \\
   & pubmed-diabetes    & 0.499\,$\pm$\,0.102 & 0.67/0.68/0.91 & 0.815\,$\pm$\,0.021 & 0.62/0.63/0.91 & 0.481\,$\pm$\,0.120 & 0.57/0.58/0.89 \\
   & wikics             & 0.078\,$\pm$\,0.010 & 0.54/0.49/0.88 & 0.734\,$\pm$\,0.013 & 0.51/0.55/0.87 & 0.097\,$\pm$\,0.012 & 0.52/0.53/0.87 \\
\midrule
  \multirow{6}{*}{TEA-GLM} & ogbn-arxiv         & 0.467\,$\pm$\,0.029 & 0.67/0.61/0.90 & 0.673\,$\pm$\,0.027 & 0.61/0.65/0.90 & 0.703\,$\pm$\,0.026 & 0.64/0.64/0.90 \\
   & ogbn-products      & 0.073\,$\pm$\,0.015 & 0.45/0.49/0.84 & 0.678\,$\pm$\,0.027 & 0.48/0.54/0.86 & 0.073\,$\pm$\,0.015 & 0.36/0.46/0.85 \\
   & patents            & 0.027\,$\pm$\,0.009 & 0.57/0.54/0.88 & 0.675\,$\pm$\,0.028 & 0.56/0.50/0.89 & 0.000\,$\pm$\,0.001 & 0.35/0.39/0.87 \\
   & physics-se         & 0.020\,$\pm$\,0.008 & 0.54/0.48/0.89 & 0.724\,$\pm$\,0.026 & 0.47/0.48/0.89 & 0.030\,$\pm$\,0.010 & 0.35/0.40/0.86 \\
   & pubmed-diabetes    & 0.457\,$\pm$\,0.029 & 0.64/0.66/0.91 & 0.845\,$\pm$\,0.022 & 0.61/0.65/0.91 & 0.470\,$\pm$\,0.029 & 0.55/0.57/0.89 \\
   & wikics             & 0.153\,$\pm$\,0.021 & 0.57/0.52/0.88 & 0.731\,$\pm$\,0.026 & 0.53/0.55/0.87 & 0.210\,$\pm$\,0.024 & 0.57/0.58/0.88 \\
\midrule
  \multirow{6}{*}{RGLM} & ogbn-arxiv         & 0.307\,$\pm$\,0.231 & 0.62/0.55/0.90 & 0.693\,$\pm$\,0.035 & 0.58/0.60/0.90 & 0.483\,$\pm$\,0.390 & 0.54/0.55/0.89 \\
   & ogbn-products      & 0.280\,$\pm$\,0.335 & 0.49/0.56/0.85 & 0.745\,$\pm$\,0.035 & 0.54/0.60/0.87 & 0.423\,$\pm$\,0.523 & 0.49/0.60/0.87 \\
   & patents            & 0.050\,$\pm$\,0.012 & 0.63/0.58/0.89 & 0.750\,$\pm$\,0.047 & 0.56/0.53/0.89 & 0.034\,$\pm$\,0.014 & 0.44/0.48/0.87 \\
   & physics-se         & 0.013\,$\pm$\,0.007 & 0.52/0.46/0.88 & 0.899\,$\pm$\,0.018 & 0.48/0.46/0.89 & 0.060\,$\pm$\,0.014 & 0.37/0.41/0.87 \\
   & pubmed-diabetes    & 0.587\,$\pm$\,0.028 & 0.76/0.74/0.93 & 0.850\,$\pm$\,0.021 & 0.64/0.61/0.91 & 0.623\,$\pm$\,0.028 & 0.63/0.61/0.90 \\
   & wikics             & 0.080\,$\pm$\,0.016 & 0.54/0.47/0.87 & 0.762\,$\pm$\,0.025 & 0.53/0.56/0.87 & 0.070\,$\pm$\,0.015 & 0.50/0.51/0.87 \\
\midrule
  \multirow{6}{*}{GOFA} & ogbn-arxiv         & 0.042\,$\pm$\,0.002 & 0.44/0.34/0.88 & 0.737\,$\pm$\,0.005 & 0.52/0.64/0.90 & 0.062\,$\pm$\,0.002 & 0.31/0.29/0.87 \\
   & ogbn-products      & 0.313\,$\pm$\,0.391 & 0.55/0.60/0.86 & 0.708\,$\pm$\,0.007 & 0.53/0.61/0.87 & 0.473\,$\pm$\,0.509 & 0.49/0.62/0.87 \\
   & patents            & 0.055\,$\pm$\,0.016 & 0.62/0.57/0.89 & 0.788\,$\pm$\,0.007 & 0.59/0.56/0.90 & 0.048\,$\pm$\,0.012 & 0.45/0.49/0.87 \\
   & physics-se         & 0.260\,$\pm$\,0.211 & 0.68/0.63/0.90 & 0.869\,$\pm$\,0.040 & 0.61/0.59/0.90 & 0.467\,$\pm$\,0.384 & 0.59/0.61/0.90 \\
   & pubmed-diabetes    & 0.675\,$\pm$\,0.262 & 0.78/0.76/0.93 & 0.877\,$\pm$\,0.009 & 0.69/0.69/0.92 & 0.707\,$\pm$\,0.372 & 0.71/0.71/0.92 \\
   & wikics             & 0.690\,$\pm$\,0.027 & 0.72/0.71/0.90 & 0.720\,$\pm$\,0.026 & 0.60/0.63/0.88 & 0.887\,$\pm$\,0.018 & 0.78/0.80/0.91 \\
\midrule
  \multirow{6}{*}{InstructGLM} & ogbn-arxiv         & 0.483\,$\pm$\,0.001 & 0.82/0.81/0.93 & 0.478\,$\pm$\,0.001 & 0.65/0.73/0.89 & 0.717\,$\pm$\,0.001 & 0.81/0.83/0.93 \\
   & ogbn-products      & 0.410\,$\pm$\,0.001 & 0.31/0.34/0.81 & 0.478\,$\pm$\,0.001 & 0.04/0.06/0.76 & 0.837\,$\pm$\,0.001 & 0.08/0.10/0.75 \\
   & patents            & 0.423\,$\pm$\,0.001 & 0.83/0.83/0.93 & 0.515\,$\pm$\,0.001 & 0.58/0.60/0.87 & 0.727\,$\pm$\,0.001 & 0.82/0.85/0.94 \\
   & physics-se         & 0.393\,$\pm$\,0.001 & -0.00/0.05/0.76 & 0.493\,$\pm$\,0.001 & -0.01/0.03/0.76 & 0.683\,$\pm$\,0.001 & 0.03/0.08/0.76 \\
   & pubmed-diabetes    & 0.810\,$\pm$\,0.001 & 0.84/0.85/0.95 & 0.519\,$\pm$\,0.001 & 0.71/0.73/0.90 & 0.937\,$\pm$\,0.001 & 0.83/0.85/0.94 \\
   & wikics             & 0.717\,$\pm$\,0.001 & 0.84/0.83/0.93 & 0.485\,$\pm$\,0.001 & 0.57/0.63/0.85 & 0.900\,$\pm$\,0.001 & 0.84/0.86/0.93 \\
\bottomrule
\end{tabular}
\renewcommand{\arraystretch}{1.0}
\end{table*}

\begin{table*}[h]
\centering
\caption{Per-domain results for every Graph-Token baseline (T4--T5). SBERT score reported across MiniLM-L6-v2 (M), mpnet-base-v2 (p), e5-large-v2 (E5) jointly. Each cell is mean$\pm$std across 3 seed.}
\label{tab:appendix_expanded_t45}
\scriptsize
\renewcommand{\arraystretch}{1.05}
\setlength{\tabcolsep}{2.5pt}
\begin{tabular}{@{}ll cc cc @{}}
\toprule
& & \multicolumn{2}{c}{\textbf{T4}} & \multicolumn{2}{c}{\textbf{T5}} \\
\cmidrule(lr){3-4} \cmidrule(lr){5-6}
\textbf{Model} & \textbf{Domain} & H@1 & SBERT (M/p/E5)  & NMI & SBERT (M/p/E5) \\
\midrule
  \multirow{6}{*}{LLaGA} & ogbn-arxiv         & 0.104\,$\pm$\,0.001 & 0.59/0.36/0.88 & 0.028\,$\pm$\,0.001 & 0.70/0.51/0.91 \\
   & ogbn-products      & 0.041\,$\pm$\,0.015 & 0.47/0.35/0.85 & 0.052\,$\pm$\,0.066 & 0.40/0.39/0.83 \\
   & patents            & 0.062\,$\pm$\,0.013 & 0.56/0.30/0.87 & 0.088\,$\pm$\,0.012 & 0.61/0.43/0.89 \\
   & physics-se         & 0.086\,$\pm$\,0.020 & 0.44/0.34/0.87 & 0.089\,$\pm$\,0.046 & 0.35/0.38/0.87 \\
   & pubmed-diabetes    & 0.066\,$\pm$\,0.051 & 0.61/0.39/0.89 & 0.051\,$\pm$\,0.007 & 0.54/0.39/0.89 \\
   & wikics             & 0.035\,$\pm$\,0.030 & 0.43/0.33/0.85 & 0.088\,$\pm$\,0.017 & 0.56/0.56/0.87 \\
\midrule
  \multirow{6}{*}{GraphToken} & ogbn-arxiv         & 0.109\,$\pm$\,0.007 & 0.46/0.34/0.87 & 0.033\,$\pm$\,0.008 & 0.57/0.41/0.90 \\
   & ogbn-products      & 0.049\,$\pm$\,0.002 & 0.38/0.48/0.83 & 0.008\,$\pm$\,0.001 & 0.44/0.46/0.84 \\
   & patents            & 0.057\,$\pm$\,0.000 & 0.55/0.28/0.87 & 0.083\,$\pm$\,0.002 & 0.64/0.42/0.90 \\
   & physics-se         & 0.240\,$\pm$\,0.006 & 0.43/0.10/0.87 & 0.069\,$\pm$\,0.005 & 0.35/0.22/0.88 \\
   & pubmed-diabetes    & 0.086\,$\pm$\,0.009 & 0.60/0.22/0.88 & 0.054\,$\pm$\,0.003 & 0.61/0.23/0.90 \\
   & wikics             & 0.066\,$\pm$\,0.005 & 0.50/0.13/0.85 & 0.038\,$\pm$\,0.007 & 0.70/0.62/0.90 \\
\midrule
  \multirow{6}{*}{GraphGPT} & ogbn-arxiv         & 0.084\,$\pm$\,0.027 & 0.34/0.31/0.87 & 0.031\,$\pm$\,0.002 & 0.35/0.35/0.88 \\
   & ogbn-products      & 0.016\,$\pm$\,0.008 & 0.35/0.37/0.83 & 0.076\,$\pm$\,0.007 & 0.39/0.40/0.82 \\
   & patents            & 0.060\,$\pm$\,0.009 & 0.54/0.25/0.86 & 0.080\,$\pm$\,0.003 & 0.63/0.40/0.90 \\
   & physics-se         & 0.251\,$\pm$\,0.001 & 0.42/0.28/0.87 & 0.079\,$\pm$\,0.000 & 0.36/0.32/0.88 \\
   & pubmed-diabetes    & 0.088\,$\pm$\,0.009 & 0.59/0.52/0.88 & 0.063\,$\pm$\,0.004 & 0.55/0.52/0.90 \\
   & wikics             & 0.063\,$\pm$\,0.007 & 0.47/0.48/0.85 & 0.035\,$\pm$\,0.004 & 0.69/0.66/0.89 \\
\midrule
  \multirow{6}{*}{TEA-GLM} & ogbn-arxiv         & 0.107\,$\pm$\,0.018 & 0.54/0.49/0.88 & 0.033\,$\pm$\,0.003 & 0.70/0.67/0.91 \\
   & ogbn-products      & 0.033\,$\pm$\,0.010 & 0.40/0.47/0.83 & 0.011\,$\pm$\,0.002 & 0.37/0.37/0.84 \\
   & patents            & 0.065\,$\pm$\,0.014 & 0.46/0.45/0.85 & 0.100\,$\pm$\,0.004 & 0.58/0.54/0.89 \\
   & physics-se         & 0.211\,$\pm$\,0.024 & 0.44/0.39/0.87 & 0.086\,$\pm$\,0.006 & 0.39/0.43/0.88 \\
   & pubmed-diabetes    & 0.106\,$\pm$\,0.018 & 0.52/0.51/0.87 & 0.061\,$\pm$\,0.004 & 0.62/0.60/0.91 \\
   & wikics             & 0.070\,$\pm$\,0.015 & 0.46/0.47/0.85 & 0.035\,$\pm$\,0.003 & 0.66/0.63/0.89 \\
\midrule
  \multirow{6}{*}{RGLM} & ogbn-arxiv         & 0.103\,$\pm$\,0.001 & 0.49/0.47/0.87 & 0.033\,$\pm$\,0.002 & 0.61/0.56/0.90 \\
   & ogbn-products      & 0.050\,$\pm$\,0.001 & 0.50/0.50/0.85 & 0.002\,$\pm$\,0.001 & 0.55/0.56/0.86 \\
   & patents            & 0.052\,$\pm$\,0.007 & 0.54/0.35/0.87 & 0.091\,$\pm$\,0.012 & 0.58/0.47/0.89 \\
   & physics-se         & 0.252\,$\pm$\,0.026 & 0.44/0.11/0.87 & 0.079\,$\pm$\,0.005 & 0.38/0.41/0.88 \\
   & pubmed-diabetes    & 0.074\,$\pm$\,0.015 & 0.62/0.08/0.89 & 0.051\,$\pm$\,0.004 & 0.62/0.12/0.90 \\
   & wikics             & 0.051\,$\pm$\,0.013 & 0.43/0.46/0.84 & 0.044\,$\pm$\,0.004 & 0.66/0.68/0.89 \\
\midrule
  \multirow{6}{*}{GOFA} & ogbn-arxiv         & 0.112\,$\pm$\,0.002 & 0.34/0.39/0.86 & 0.028\,$\pm$\,0.005 & 0.46/0.42/0.89 \\
   & ogbn-products      & 0.048\,$\pm$\,0.002 & 0.49/0.50/0.85 & 0.006\,$\pm$\,0.002 & 0.55/0.54/0.86 \\
   & patents            & 0.055\,$\pm$\,0.002 & 0.56/0.33/0.87 & 0.075\,$\pm$\,0.002 & 0.67/0.50/0.91 \\
   & physics-se         & 0.251\,$\pm$\,0.013 & 0.57/0.24/0.88 & 0.068\,$\pm$\,0.006 & 0.62/0.48/0.90 \\
   & pubmed-diabetes    & 0.083\,$\pm$\,0.009 & 0.65/0.43/0.89 & 0.051\,$\pm$\,0.001 & 0.70/0.21/0.91 \\
   & wikics             & 0.063\,$\pm$\,0.014 & 0.59/0.50/0.87 & 0.025\,$\pm$\,0.004 & 0.78/0.74/0.91 \\
\midrule
  \multirow{6}{*}{InstructGLM} & ogbn-arxiv         & 0.066\,$\pm$\,0.001 & 0.55/0.51/0.87 & 0.023\,$\pm$\,0.001 & 0.18/0.13/0.76 \\
   & ogbn-products      & 0.025\,$\pm$\,0.001 & 0.03/0.07/0.75 & 0.002\,$\pm$\,0.001 & 0.15/0.13/0.75 \\
   & patents            & 0.085\,$\pm$\,0.001 & 0.29/0.21/0.79 & 0.054\,$\pm$\,0.001 & 0.19/0.13/0.75 \\
   & physics-se         & 0.143\,$\pm$\,0.001 & 0.01/0.07/0.77 & 0.053\,$\pm$\,0.001 & 0.09/0.23/0.81 \\
   & pubmed-diabetes    & 0.140\,$\pm$\,0.001 & 0.43/0.38/0.84 & 0.021\,$\pm$\,0.001 & 0.07/-0.01/0.74 \\
   & wikics             & 0.018\,$\pm$\,0.001 & 0.66/0.69/0.89 & 0.007\,$\pm$\,0.001 & 0.46/0.48/0.83 \\
\bottomrule
\end{tabular}
\renewcommand{\arraystretch}{1.0}
\end{table*}


\section{Per-Domain Results: Zero-shot LLMs}
\label{app:percell_zeroshot}
Per-(domain, task) hard accuracy and 3-SBERT triples for closed-source
zero-shot LLMs (Claude Opus 4.7, GPT-5). Sampling: $n{=}20$ test samples
per (domain, task) cell, fixed seed=0, 6 domains $\times$ 5 tasks. Prompt is
format-only (canonical sentence frame, no closed-set label list shown). For
T1/T3, hard accuracy is computed after projecting each free-form Answer
onto the nearest canonical label using SBERT-MiniLM cosine over the
domain-task label set.

\begin{table*}[h]
\centering
\caption{Per-domain results for zero-shot closed-source LLMs (T1--T3),
n{=}20 samples per (domain, task) cell, format-only prompt with
SBERT-MiniLM nearest-canonical-label projection for T1/T3
(Section~\ref{sec:abl_reasoning}). SBERT score reported across MiniLM-L6-v2
(M), mpnet-base-v2 (p), e5-large-v2 (E5) jointly.}
\label{tab:appendix_zeroshot_t123}
\scriptsize
\renewcommand{\arraystretch}{1.05}
\setlength{\tabcolsep}{2.5pt}
\begin{tabular}{@{}ll cc cc cc @{}}
\toprule
& & \multicolumn{2}{c}{\textbf{T1}} & \multicolumn{2}{c}{\textbf{T2}} & \multicolumn{2}{c}{\textbf{T3}} \\
\cmidrule(lr){3-4} \cmidrule(lr){5-6} \cmidrule(lr){7-8}
\textbf{Model} & \textbf{Domain} & Acc & SBERT (M/p/E5)  & Acc & SBERT (M/p/E5)  & Acc & SBERT (M/p/E5) \\
\midrule
  \multirow{6}{*}{GPT-5 (zero-shot)} & ogbn-arxiv         & 0.200 & 0.68/0.75/0.89 & 0.850 & 0.75/0.75/0.93 & 0.600 & 0.76/0.79/0.90 \\
   & ogbn-products      & 0.300 & 0.77/0.80/0.92 & 0.700 & 0.78/0.73/0.94 & 0.500 & 0.70/0.75/0.91 \\
   & pubmed-diabetes    & 0.800 & 0.78/0.77/0.92 & 0.700 & 0.82/0.79/0.95 & 0.850 & 0.82/0.87/0.92 \\
   & wikics             & 0.650 & 0.78/0.77/0.91 & 0.800 & 0.73/0.73/0.92 & 1.000 & 0.80/0.81/0.91 \\
   & patents            & 0.250 & 0.71/0.76/0.91 & 1.000 & 0.74/0.71/0.92 & 0.550 & 0.76/0.79/0.91 \\
   & physics-se         & 0.400 & 0.65/0.73/0.89 & 0.650 & 0.81/0.76/0.93 & 0.700 & 0.75/0.77/0.90 \\
\midrule
  \multirow{6}{*}{Claude Opus 4.7 (zero-shot)} & ogbn-arxiv         & 0.400 & 0.79/0.81/0.93 & 1.000 & 0.79/0.74/0.94 & 0.500 & 0.78/0.82/0.90 \\
   & ogbn-products      & 0.250 & 0.70/0.72/0.90 & 0.950 & 0.83/0.73/0.95 & 0.500 & 0.71/0.71/0.90 \\
   & pubmed-diabetes    & 0.700 & 0.80/0.80/0.94 & 1.000 & 0.83/0.77/0.95 & 0.850 & 0.84/0.88/0.93 \\
   & wikics             & 0.850 & 0.68/0.71/0.89 & 1.000 & 0.75/0.71/0.93 & 1.000 & 0.79/0.82/0.91 \\
   & patents            & 0.200 & 0.80/0.80/0.93 & 1.000 & 0.87/0.82/0.95 & 0.600 & 0.80/0.84/0.92 \\
   & physics-se         & 0.400 & 0.84/0.80/0.94 & 1.000 & 0.82/0.75/0.94 & 0.750 & 0.75/0.76/0.91 \\
\bottomrule
\end{tabular}
\renewcommand{\arraystretch}{1.0}
\end{table*}

\begin{table*}[h]
\centering
\caption{Per-domain results for zero-shot closed-source LLMs (T4--T5),
n{=}20 samples per (domain, task) cell. SBERT score reported across
MiniLM-L6-v2 (M), mpnet-base-v2 (p), e5-large-v2 (E5) jointly.}
\label{tab:appendix_zeroshot_t45}
\scriptsize
\renewcommand{\arraystretch}{1.05}
\setlength{\tabcolsep}{2.5pt}
\begin{tabular}{@{}ll cc cc @{}}
\toprule
& & \multicolumn{2}{c}{\textbf{T4}} & \multicolumn{2}{c}{\textbf{T5}} \\
\cmidrule(lr){3-4} \cmidrule(lr){5-6}
\textbf{Model} & \textbf{Domain} & H@1 & SBERT (M/p/E5)  & NMI & SBERT (M/p/E5) \\
\midrule
  \multirow{6}{*}{GPT-5 (zero-shot)} & ogbn-arxiv         & 0.550 & 0.76/0.72/0.91 & 0.219 & 0.80/0.82/0.94 \\
   & ogbn-products      & 0.300 & 0.66/0.68/0.89 & 0.223 & 0.82/0.83/0.92 \\
   & pubmed-diabetes    & 0.400 & 0.72/0.71/0.90 & 0.353 & 0.84/0.88/0.94 \\
   & wikics             & 0.450 & 0.64/0.66/0.88 & 0.347 & 0.82/0.84/0.92 \\
   & patents            & 0.350 & 0.69/0.71/0.89 & 0.228 & 0.78/0.84/0.93 \\
   & physics-se         & 0.700 & 0.74/0.75/0.92 & 0.191 & 0.83/0.85/0.94 \\
\midrule
  \multirow{6}{*}{Claude Opus 4.7 (zero-shot)} & ogbn-arxiv         & 0.650 & 0.69/0.63/0.90 & 0.327 & 0.64/0.57/0.88 \\
   & ogbn-products      & 0.700 & 0.66/0.70/0.90 & 0.251 & 0.80/0.80/0.91 \\
   & pubmed-diabetes    & 0.500 & 0.69/0.69/0.91 & 0.357 & 0.85/0.86/0.95 \\
   & wikics             & 0.850 & 0.72/0.69/0.90 & 0.432 & 0.80/0.80/0.92 \\
   & patents            & 0.450 & 0.67/0.67/0.90 & 0.301 & 0.62/0.62/0.89 \\
   & physics-se         & 0.700 & 0.77/0.77/0.92 & 0.200 & 0.81/0.83/0.94 \\
\bottomrule
\end{tabular}
\renewcommand{\arraystretch}{1.0}
\end{table*}

\section{Per-Domain Results: SFT (Graph2Text) Baselines}
\label{app:percell_sft}
Per-(domain, task) hard accuracy and 3-SBERT triples for the three
Graph2Text SFT baselines (Vicuna-7B, Qwen2-7B, Llama-3-8B), fine-tuned
end-to-end with QLoRA on the per-domain training split.

\begin{table*}[h]
\centering
\caption{Per-domain results for SFT baselines (T1--T3). SBERT score across MiniLM-L6-v2 (M), mpnet-base-v2 (p), e5-large-v2 (E5) jointly. Each cell is mean$\pm$std across 3 seed.}
\label{tab:appendix_sft_t123}
\scriptsize
\renewcommand{\arraystretch}{1.05}
\setlength{\tabcolsep}{2.5pt}
\begin{tabular}{@{}ll cc cc cc @{}}
\toprule
& & \multicolumn{2}{c}{\textbf{T1}} & \multicolumn{2}{c}{\textbf{T2}} & \multicolumn{2}{c}{\textbf{T3}} \\
\cmidrule(lr){3-4} \cmidrule(lr){5-6} \cmidrule(lr){7-8}
\textbf{Model} & \textbf{Domain} & Acc & SBERT (M/p/E5)  & Acc & SBERT (M/p/E5)  & Acc & SBERT (M/p/E5) \\
\midrule
  \multirow{6}{*}{SFT (Vicuna-7B)} & ogbn-arxiv         & 0.415\,$\pm$\,0.051 & 0.81/0.81/0.93 & 0.577\,$\pm$\,0.042 & 0.86/0.85/0.94 & 0.607\,$\pm$\,0.166 & 0.79/0.81/0.93 \\
   & ogbn-products      & 0.682\,$\pm$\,0.001 & 0.63/0.72/0.90 & 0.708\,$\pm$\,0.001 & 0.79/0.77/0.93 & 0.726\,$\pm$\,0.001 & 0.74/0.81/0.92 \\
   & patents            & 0.426\,$\pm$\,0.001 & 0.83/0.84/0.94 & 0.732\,$\pm$\,0.001 & 0.83/0.80/0.95 & 0.746\,$\pm$\,0.001 & 0.84/0.87/0.94 \\
   & physics-se         & 0.413\,$\pm$\,0.001 & 0.87/0.83/0.95 & 0.897\,$\pm$\,0.001 & 0.88/0.83/0.96 & 0.680\,$\pm$\,0.001 & 0.83/0.83/0.94 \\
   & pubmed-diabetes    & 0.843\,$\pm$\,0.052 & 0.87/0.86/0.95 & 0.793\,$\pm$\,0.014 & 0.88/0.86/0.96 & 0.974\,$\pm$\,0.026 & 0.86/0.87/0.94 \\
   & wikics             & 0.700\,$\pm$\,0.001 & 0.79/0.79/0.92 & 0.776\,$\pm$\,0.001 & 0.76/0.76/0.92 & 0.783\,$\pm$\,0.001 & 0.87/0.89/0.94 \\
\midrule
  \multirow{6}{*}{SFT (Qwen2-7B)} & ogbn-arxiv         & 0.440\,$\pm$\,0.014 & 0.84/0.84/0.94 & 0.784\,$\pm$\,0.007 & 0.87/0.86/0.95 & 0.700\,$\pm$\,0.028 & 0.82/0.85/0.94 \\
   & ogbn-products      & 0.366\,$\pm$\,0.001 & 0.72/0.79/0.91 & 0.703\,$\pm$\,0.001 & 0.75/0.74/0.93 & 0.636\,$\pm$\,0.001 & 0.76/0.84/0.93 \\
   & patents            & 0.325\,$\pm$\,0.040 & 0.83/0.84/0.94 & 0.828\,$\pm$\,0.001 & 0.87/0.84/0.96 & 0.552\,$\pm$\,0.049 & 0.84/0.87/0.94 \\
   & physics-se         & 0.200\,$\pm$\,0.001 & 0.86/0.81/0.94 & 0.793\,$\pm$\,0.001 & 0.86/0.80/0.95 & 0.413\,$\pm$\,0.001 & 0.81/0.82/0.93 \\
   & pubmed-diabetes    & 0.870\,$\pm$\,0.001 & 0.87/0.87/0.96 & 0.890\,$\pm$\,0.001 & 0.89/0.87/0.96 & 0.983\,$\pm$\,0.001 & 0.87/0.89/0.95 \\
   & wikics             & 0.686\,$\pm$\,0.001 & 0.84/0.85/0.93 & 0.732\,$\pm$\,0.001 & 0.84/0.81/0.93 & 0.905\,$\pm$\,0.001 & 0.89/0.90/0.95 \\
\midrule
  \multirow{6}{*}{SFT (Llama-3-8B)} & ogbn-arxiv         & 0.467\,$\pm$\,0.015 & 0.84/0.84/0.94 & 0.854\,$\pm$\,0.025 & 0.89/0.88/0.96 & 0.713\,$\pm$\,0.007 & 0.84/0.86/0.94 \\
   & ogbn-products      & 0.503\,$\pm$\,0.051 & 0.75/0.81/0.92 & 0.758\,$\pm$\,0.079 & 0.84/0.80/0.95 & 0.684\,$\pm$\,0.039 & 0.76/0.84/0.93 \\
   & patents            & 0.366\,$\pm$\,0.027 & 0.84/0.85/0.94 & 0.857\,$\pm$\,0.012 & 0.88/0.85/0.96 & 0.593\,$\pm$\,0.028 & 0.82/0.87/0.94 \\
   & physics-se         & 0.417\,$\pm$\,0.001 & 0.87/0.83/0.95 & 0.722\,$\pm$\,0.049 & 0.89/0.83/0.96 & 0.690\,$\pm$\,0.005 & 0.83/0.83/0.94 \\
   & pubmed-diabetes    & 0.859\,$\pm$\,0.002 & 0.88/0.88/0.96 & 0.887\,$\pm$\,0.005 & 0.90/0.87/0.97 & 0.964\,$\pm$\,0.021 & 0.87/0.88/0.95 \\
   & wikics             & 0.688\,$\pm$\,0.001 & 0.86/0.86/0.94 & 0.851\,$\pm$\,0.001 & 0.87/0.83/0.95 & 0.949\,$\pm$\,0.001 & 0.89/0.91/0.95 \\
\bottomrule
\end{tabular}
\renewcommand{\arraystretch}{1.0}
\end{table*}

\begin{table*}[h]
\centering
\caption{Per-domain results for SFT baselines (T4--T5). SBERT score across MiniLM-L6-v2 (M), mpnet-base-v2 (p), e5-large-v2 (E5) jointly. Each cell is mean$\pm$std across 3 seed.}
\label{tab:appendix_sft_t45}
\scriptsize
\renewcommand{\arraystretch}{1.05}
\setlength{\tabcolsep}{2.5pt}
\begin{tabular}{@{}ll cc cc @{}}
\toprule
& & \multicolumn{2}{c}{\textbf{T4}} & \multicolumn{2}{c}{\textbf{T5}} \\
\cmidrule(lr){3-4} \cmidrule(lr){5-6}
\textbf{Model} & \textbf{Domain} & H@1 & SBERT (M/p/E5)  & NMI & SBERT (M/p/E5) \\
\midrule
  \multirow{6}{*}{SFT (Vicuna-7B)} & ogbn-arxiv         & 0.311\,$\pm$\,0.006 & 0.70/0.68/0.91 & 0.158\,$\pm$\,0.096 & 0.56/0.56/0.87 \\
   & ogbn-products      & 0.124\,$\pm$\,0.001 & 0.48/0.50/0.84 & 0.074\,$\pm$\,0.001 & 0.31/0.34/0.78 \\
   & patents            & 0.215\,$\pm$\,0.001 & 0.68/0.67/0.90 & 0.170\,$\pm$\,0.001 & 0.46/0.44/0.84 \\
   & physics-se         & 0.403\,$\pm$\,0.001 & 0.77/0.76/0.93 & 0.106\,$\pm$\,0.001 & 0.83/0.86/0.95 \\
   & pubmed-diabetes    & 0.204\,$\pm$\,0.048 & 0.61/0.60/0.89 & 0.201\,$\pm$\,0.009 & 0.59/0.57/0.88 \\
   & wikics             & 0.055\,$\pm$\,0.001 & 0.60/0.63/0.86 & 0.069\,$\pm$\,0.001 & 0.56/0.58/0.86 \\
\midrule
  \multirow{6}{*}{SFT (Qwen2-7B)} & ogbn-arxiv         & 0.323\,$\pm$\,0.014 & 0.72/0.70/0.91 & 0.147\,$\pm$\,0.025 & 0.36/0.34/0.82 \\
   & ogbn-products      & 0.237\,$\pm$\,0.001 & 0.44/0.47/0.83 & 0.166\,$\pm$\,0.001 & 0.21/0.24/0.75 \\
   & patents            & 0.204\,$\pm$\,0.001 & 0.70/0.70/0.91 & 0.225\,$\pm$\,0.008 & 0.43/0.40/0.83 \\
   & physics-se         & 0.320\,$\pm$\,0.001 & 0.76/0.76/0.93 & 0.128\,$\pm$\,0.001 & 0.83/0.86/0.95 \\
   & pubmed-diabetes    & 0.187\,$\pm$\,0.001 & 0.64/0.63/0.89 & 0.266\,$\pm$\,0.001 & 0.58/0.56/0.88 \\
   & wikics             & 0.259\,$\pm$\,0.001 & 0.70/0.71/0.90 & 0.394\,$\pm$\,0.001 & 0.34/0.36/0.80 \\
\midrule
  \multirow{6}{*}{SFT (Llama-3-8B)} & ogbn-arxiv         & 0.456\,$\pm$\,0.013 & 0.75/0.74/0.92 & 0.185\,$\pm$\,0.006 & 0.38/0.36/0.82 \\
   & ogbn-products      & 0.315\,$\pm$\,0.007 & 0.48/0.50/0.84 & 0.198\,$\pm$\,0.011 & 0.21/0.23/0.75 \\
   & patents            & 0.273\,$\pm$\,0.001 & 0.72/0.70/0.91 & 0.254\,$\pm$\,0.022 & 0.45/0.42/0.83 \\
   & physics-se         & 0.550\,$\pm$\,0.014 & 0.81/0.79/0.94 & 0.165\,$\pm$\,0.005 & 0.86/0.89/0.96 \\
   & pubmed-diabetes    & 0.376\,$\pm$\,0.035 & 0.68/0.67/0.90 & 0.429\,$\pm$\,0.010 & 0.63/0.61/0.89 \\
   & wikics             & 0.373\,$\pm$\,0.001 & 0.76/0.76/0.92 & 0.427\,$\pm$\,0.001 & 0.34/0.36/0.80 \\
\bottomrule
\end{tabular}
\renewcommand{\arraystretch}{1.0}
\end{table*}

\end{document}